%% file: main.tex
\documentclass[lettersize,journal]{IEEEtran}

\usepackage{cite}
\usepackage{amsmath,amsfonts}
\usepackage{algorithmic}
\usepackage{algorithm}
\usepackage{array}
\usepackage[caption=false,font=normalsize,labelfont=sf,textfont=sf]{subfig}
\usepackage{textcomp}
\usepackage{stfloats}
\usepackage{url}
\usepackage{verbatim}
\usepackage{graphicx}

\usepackage[colorlinks,
            linkcolor=blue,
            anchorcolor=blue,
            citecolor=blue]{hyperref}

\usepackage{graphicx}
\usepackage{amssymb}
\usepackage{booktabs}
\usepackage{color}
\usepackage{colortbl} 
\usepackage{svg}
\usepackage{float}
\usepackage{subcaption}
\usepackage{multirow}
\usepackage{amssymb} 
\usepackage{makecell}
\usepackage{bm}
\usepackage{float}

\usepackage{marvosym}

\begin{document}

\title{Vchitect-2.0: Parallel Transformer for Scaling Up Video Diffusion Models}

\author{Weichen~Fan$^\dag$,
        Chenyang~Si$^\dag$,
        Junhao Song,
        Zhenyu~Yang,
        Yinan He,
        Long Zhuo,
        Ziqi Huang,
        Ziyue Dong,
        Jingwen He,
        Dongwei Pan,
        Yi Wang,
        Yuming Jiang,
        Yaohui Wang,
        Peng Gao,
        Xinyuan Chen,
        Hengjie Li, \\
        Dahua Lin~\textsuperscript{\Letter},
        Yu Qiao~\textsuperscript{\Letter},
        and~Ziwei~Liu~\textsuperscript{\Letter}

\thanks{Weichen Fan, Chenyang Si, Ziqi Huang, Yuming Jiang, Ziwei Liu are with the S-Lab, Nanyang Technological University, Singapore, 639798. (e-mail:fanweichen2383@gmail.com, chenyang.si@ntu.edu.sg, ziwei.liu@ntu.edu.sg.)}
\thanks{Junhao Song, Zhenyu Yang, Yinan He, Long Zhuo, Ziyue Dong, Jingwen He, Dongwei Pan, Yi Wang, Yaohui Wang, Xinyuan Chen, Peng Gao, Hengjie Li, Yu Qiao are with Shanghai Artificial Intelligence Laboratory, China. }
\thanks{Dahua Lin is with The Chinese University of Hong Kong and Shanghai Artificial Intelligence Laboratory, China. }
\thanks{ $\dag$ Equal contribution.}}

\markboth{Journal of \LaTeX\ Class Files,~Vol.~14, No.~8, August~2021}%
{Shell \MakeLowercase{\textit{et al.}}: A Sample Article Using IEEEtran.cls for IEEE Journals}


\maketitle

\begin{abstract}
    We present \textbf{Vchitect-2.0}, a parallel transformer architecture designed to scale up video diffusion models for large-scale text-to-video generation. The overall Vchitect-2.0 system has several key designs. \textbf{1)} By introducing a novel \textbf{Multimodal Diffusion Block}, our approach achieves consistent alignment between text descriptions and generated video frames, while maintaining temporal coherence across sequences. \textbf{2)} To overcome memory and computational bottlenecks, we propose a \textbf{Memory-efficient Training} framework that incorporates hybrid parallelism and other memory reduction techniques, enabling efficient training of long video sequences on distributed systems. \textbf{3)} Additionally, our enhanced data processing pipeline ensures the creation of \textbf{Vchitect T2V DataVerse}, a high-quality million-scale training dataset through rigorous annotation and aesthetic evaluation. Extensive benchmarking demonstrates that Vchitect-2.0 outperforms existing methods in video quality, training efficiency, and scalability, serving as a suitable base for high-fidelity video generation.
\end{abstract}




\input{sec/1_introduction}
\input{sec/2_related_work}
\input{sec/3_method}

\input{sec/4_experiments}

\input{sec/5_conclusion}

\ifCLASSOPTIONcompsoc
  \section*{Acknowledgments}
\else
  \section*{Acknowledgment}
\fi


This study is supported by the National Key R\&D Program of China No.2022ZD0160102, and by the video generation project (Intern-Vchitect) of Shanghai Artificial Intelligence Laboratory. This study is also supported by the Ministry of Education, Singapore, under its MOE AcRF Tier 2 (MOET2EP20221- 0012), NTU NAP, and under the RIE2020 Industry Alignment Fund – Industry Collaboration Projects (IAF-ICP) Funding Initiative, as well as cash and in-kind contribution from the industry partner(s).

\ifCLASSOPTIONcaptionsoff
  \newpage
\fi



%
{ \small
  \bibliographystyle{IEEEtran}
  \bibliography{main}
}




%

\begin{IEEEbiography}[{\includegraphics[width=1in,height=1.25in,clip,keepaspectratio]{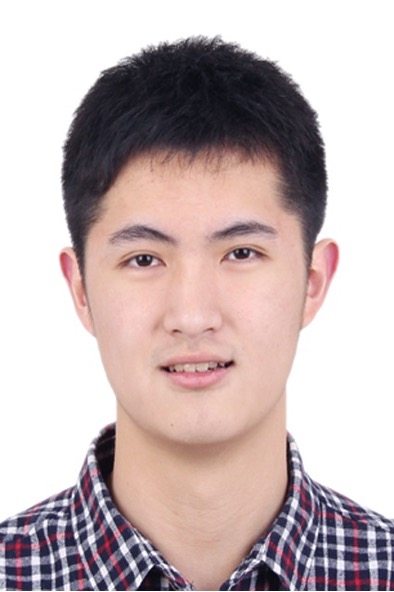}}]{Weichen Fan}
received the bachelor's degree from University of Electronic Science and Technology of China (UESTC), and the master's degree from National University of Singapore (NUS). He is currently working toward the Ph.D. degree with MMLab@NTU, Nanyang Technological University, supervised by Prof. Ziwei Liu.
His research interests include image/video generation, robotics, and robotic simulation.
\end{IEEEbiography}

\begin{IEEEbiography}[{\includegraphics[width=1in,height=1.25in,clip,keepaspectratio]{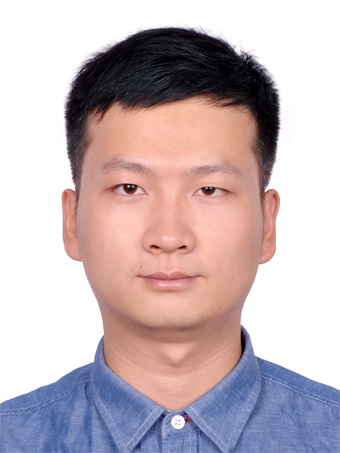}}]{Chenyang Si}
received the B.S. degree from Zhengzhou University, Zhengzhou, China, in 2016, and the Ph.D. degree from the National Laboratory of Pattern Recognition (NLPR), Institute of Automation, Chinese Academy of Sciences (CASIA), Beijing, China, in 2021. Currently, he is a research fellow at Nanyang Technological University (NTU) Singapore. His research lies at the intersection of deep learning and computer vision, including vision-based human perception (pose and action), few-shot learning, self-supervised learning, semi-supervised learning, network architecture, and image/video generation.
\end{IEEEbiography}

\begin{IEEEbiography}[{\includegraphics[width=1in,height=1.25in,clip,keepaspectratio]{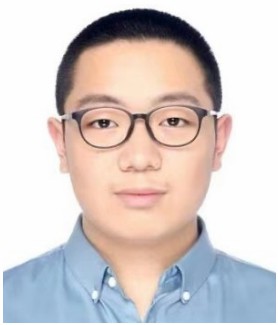}}]{Junhao Song}
received the bachelor's degree from Tsinghua University. He is currently working toward the PhD degree at Shanghai Jiao Tong University, supervised by Prof. Hengjie Li. His research interests include distributed systems for AI, image/video generation, and domain specific language.
\end{IEEEbiography}

\begin{IEEEbiography}[{\includegraphics[width=1in,height=1.25in,clip,keepaspectratio]{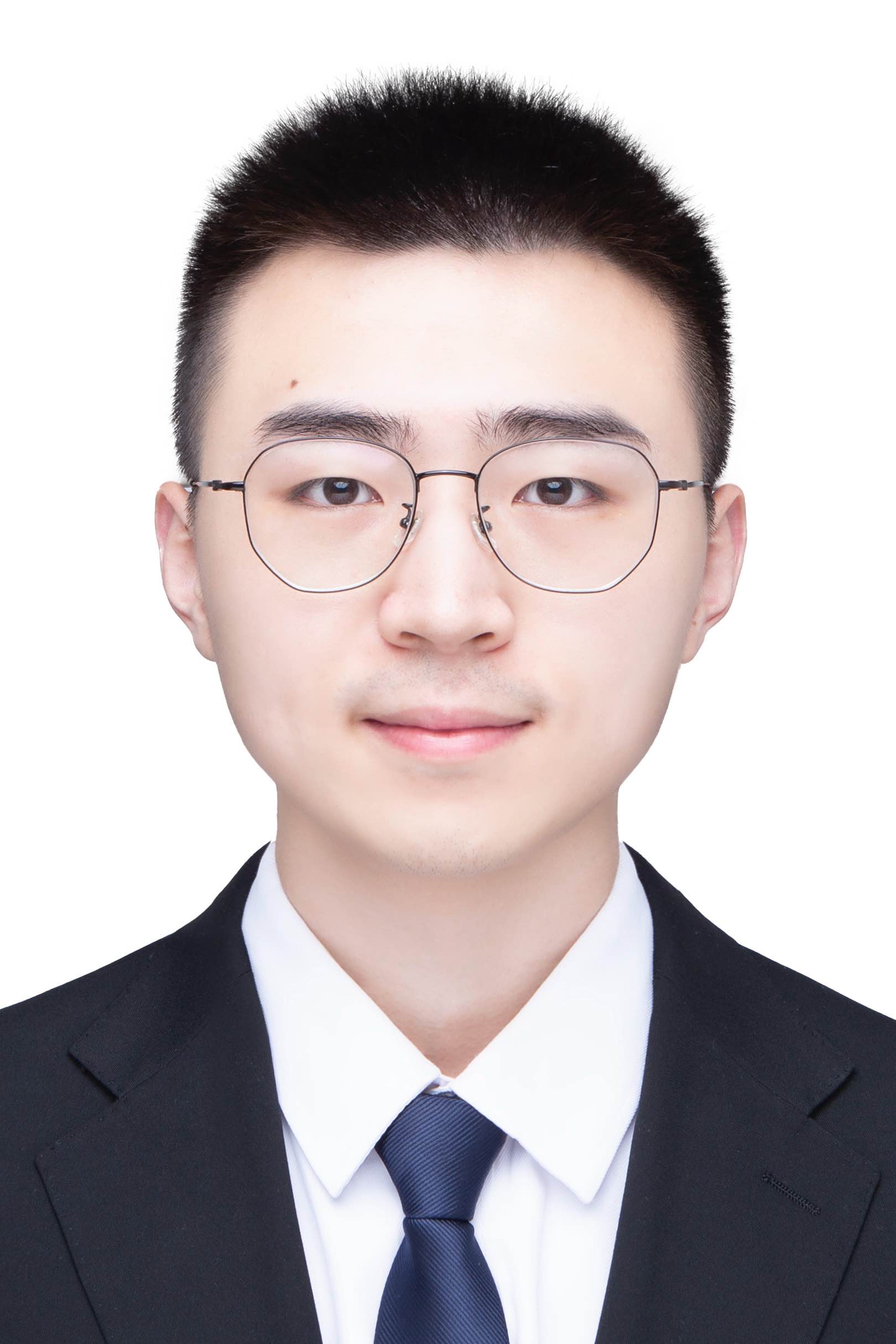}}]{Zhenyu Yang}
received the bachelor's degree from Nanjing University of Science and Technology, Nanjing, China, in 2019, and the master's degree from University of Chinese Academy of Sciences, Beijing, China, in 2022. He is currently working  at Shanghai Artificial Intelligence Laboratory. His research interests include machine learning system, video generation, and embodied artificial intelligence.
\end{IEEEbiography}

\begin{IEEEbiography}[{\includegraphics[width=1in,height=1.25in,clip,keepaspectratio]{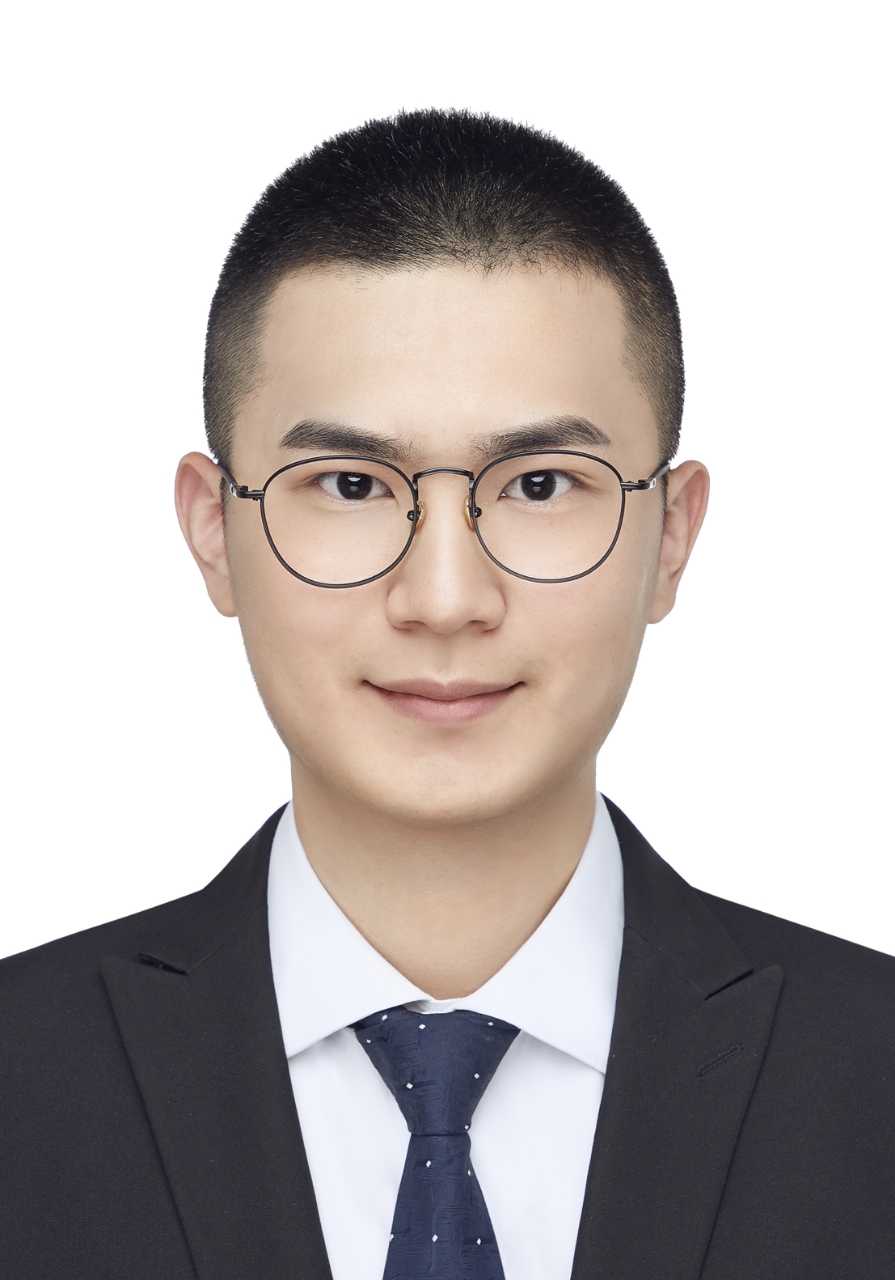}}]{Yinan He}
is currently a Research Engineer at Shanghai AI Laboratory, where he is a member of the OpenGVLab. He received his Master's degree from Beijing University of Posts and Telecommunications. His current research interests include video understanding and multi-modality large language models.
\end{IEEEbiography}

\begin{IEEEbiography}[{\includegraphics[width=1in,height=1.25in,clip,keepaspectratio]{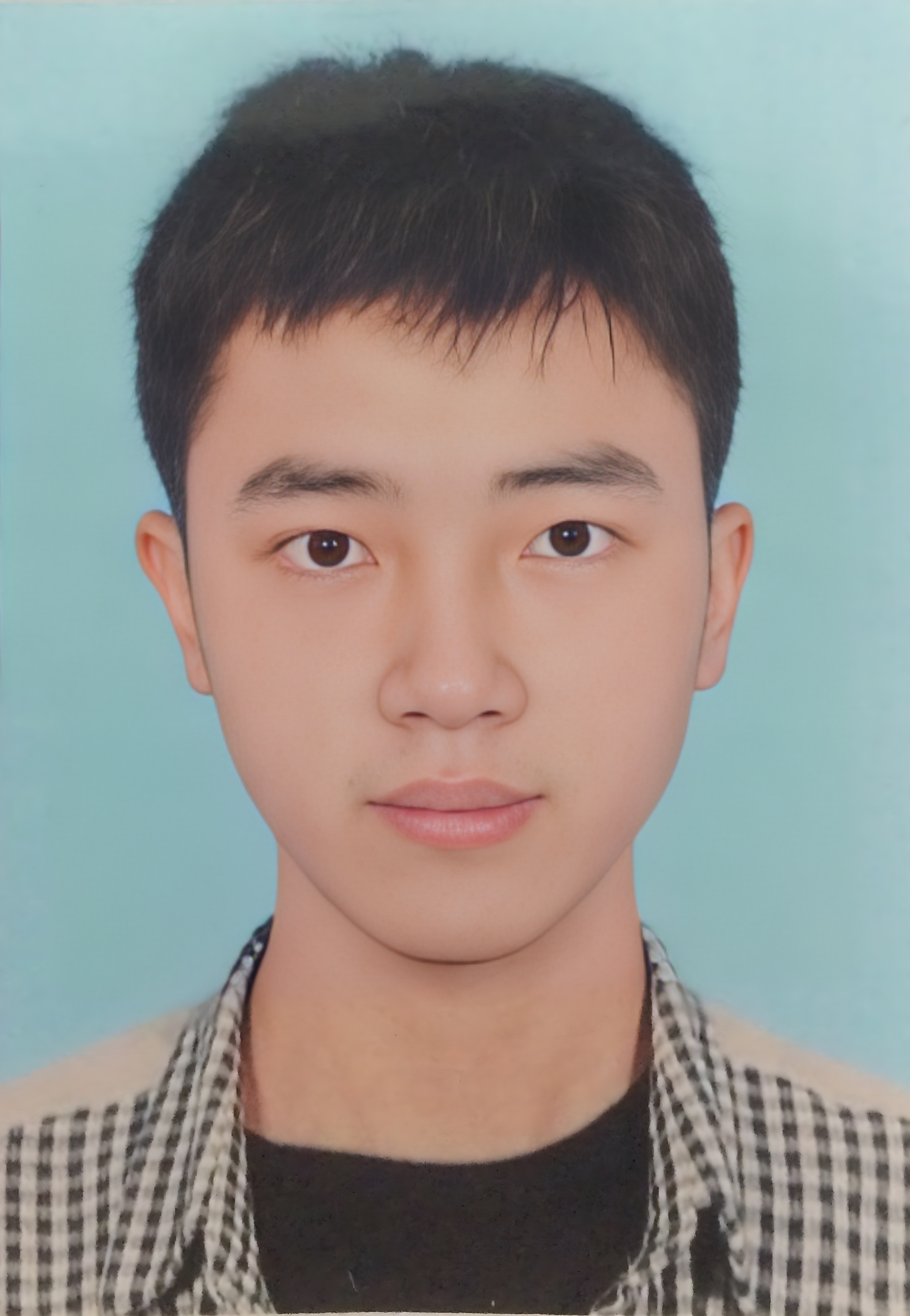}}]{Long Zhuo}
 is currently a research engineer at the Shanghai AI Laboratory. His primary research interests lie in the fields of image generation, video generation, and 3D/4D generation. Before joining the Shanghai AI Laboratory, he was a research assistant at the Shenzhen Key Laboratory of Media Security. He has published several academic papers in related fields, such as TPAMI, NeurIPS, ECCV and TIFS. 

\end{IEEEbiography}

\begin{IEEEbiography}
[{\includegraphics[width=1in,height=1.25in,clip,keepaspectratio]{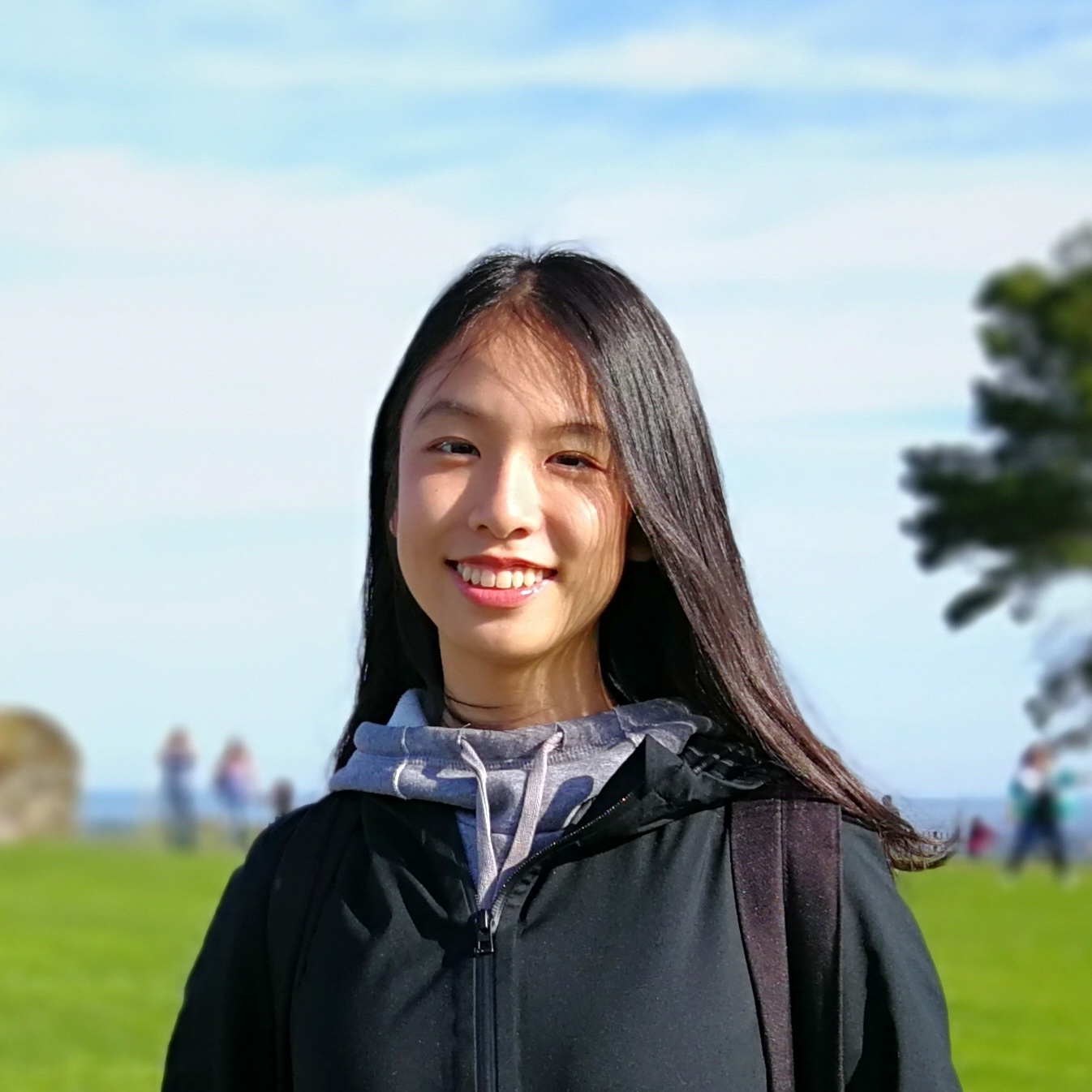}}]{Ziqi Huang} is currently a Ph.D. student at MMLab@NTU, Nanyang Technological University (NTU), supervised by Prof. Ziwei Liu. She received her Bachelor's degree from NTU in 2022. Her current research interests include visual generation and evaluation. She is awarded Google PhD Fellowship 2023.
\end{IEEEbiography}

\begin{IEEEbiography}[{\includegraphics[width=1in,height=1.25in,clip,keepaspectratio]{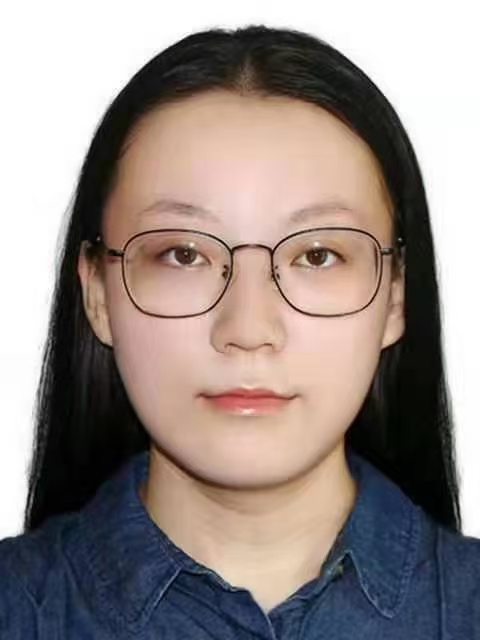}}]{Ziyue Dong} received the B.E. degree in Software Engineering from Dalian University of Technology, Dalian, China, in 2022. She is currently pursuing an M.S. degree at the School of Software Engineering, Xi'an Jiaotong University, Xi'an, China. Her research interests include image super-resolution, image generation, and video generation.
\end{IEEEbiography}

\begin{IEEEbiography}[{\includegraphics[width=1in,height=1.25in,clip,keepaspectratio]{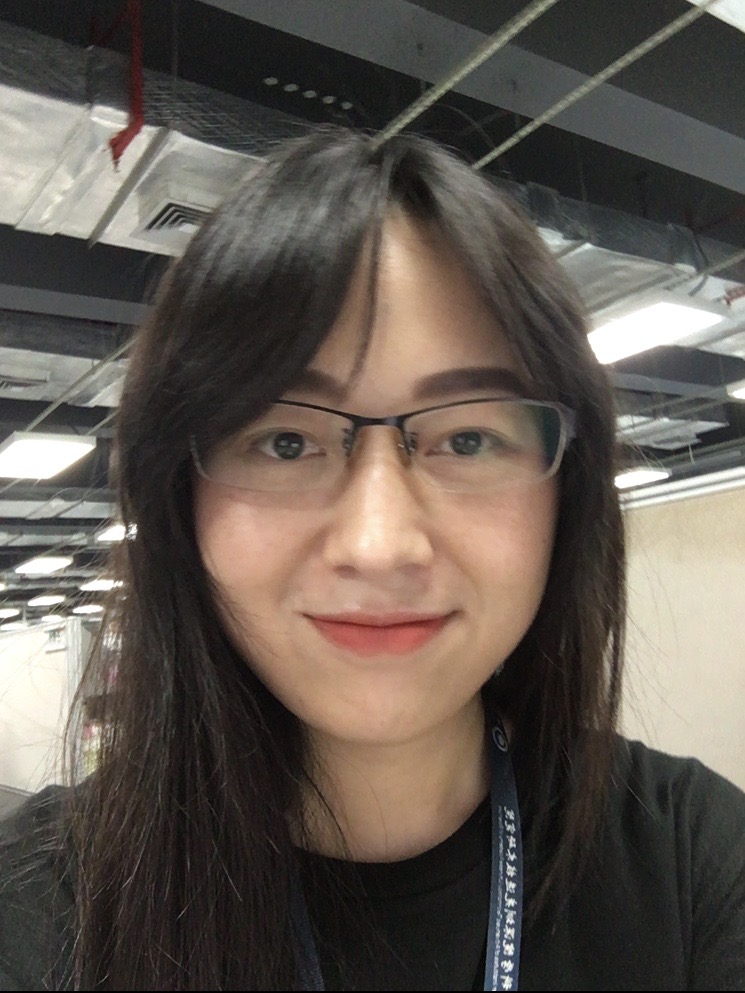}}]{Jingwen He}
received the B.Eng. degree in computer science and technology from Sichuan University, China, in 2016, and the M.Phil. degree in electronic and information engineering from the University of Sydney, Australia, in 2019. She is currently a Ph.D student with the Information Engineering department, Chinese University of Hong Kong, supervised by Prof, Wanli Ouyang. Her current research interests include image/video generation and image/video processing.
\end{IEEEbiography}

\begin{IEEEbiography}[{\includegraphics[width=1in,height=1.25in,clip,keepaspectratio]{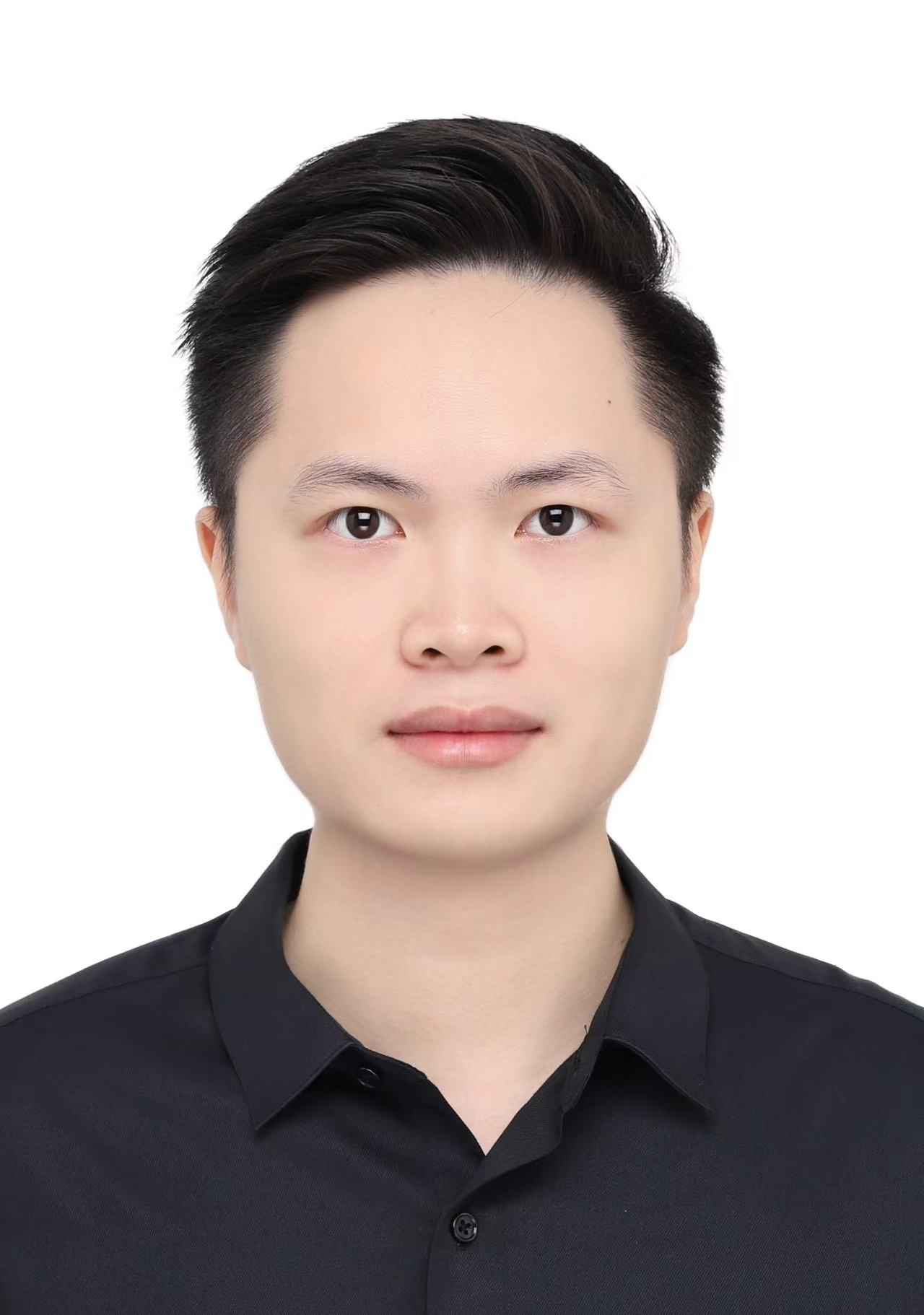}}]{Dongwei Pan}
is currently an Algorithm Engineer with the Shanghai AI Laboratory, China. He obtained his Master's degree in Karlsruhe Institute of Technology, Germany. His research interests include human reconstruction, human generation, video generation and robotics. He has published several papers, including NeurIPS, CVPR and ROBIO.
\end{IEEEbiography}

\begin{IEEEbiography}[{\includegraphics[width=1in,height=1.25in,clip,keepaspectratio]{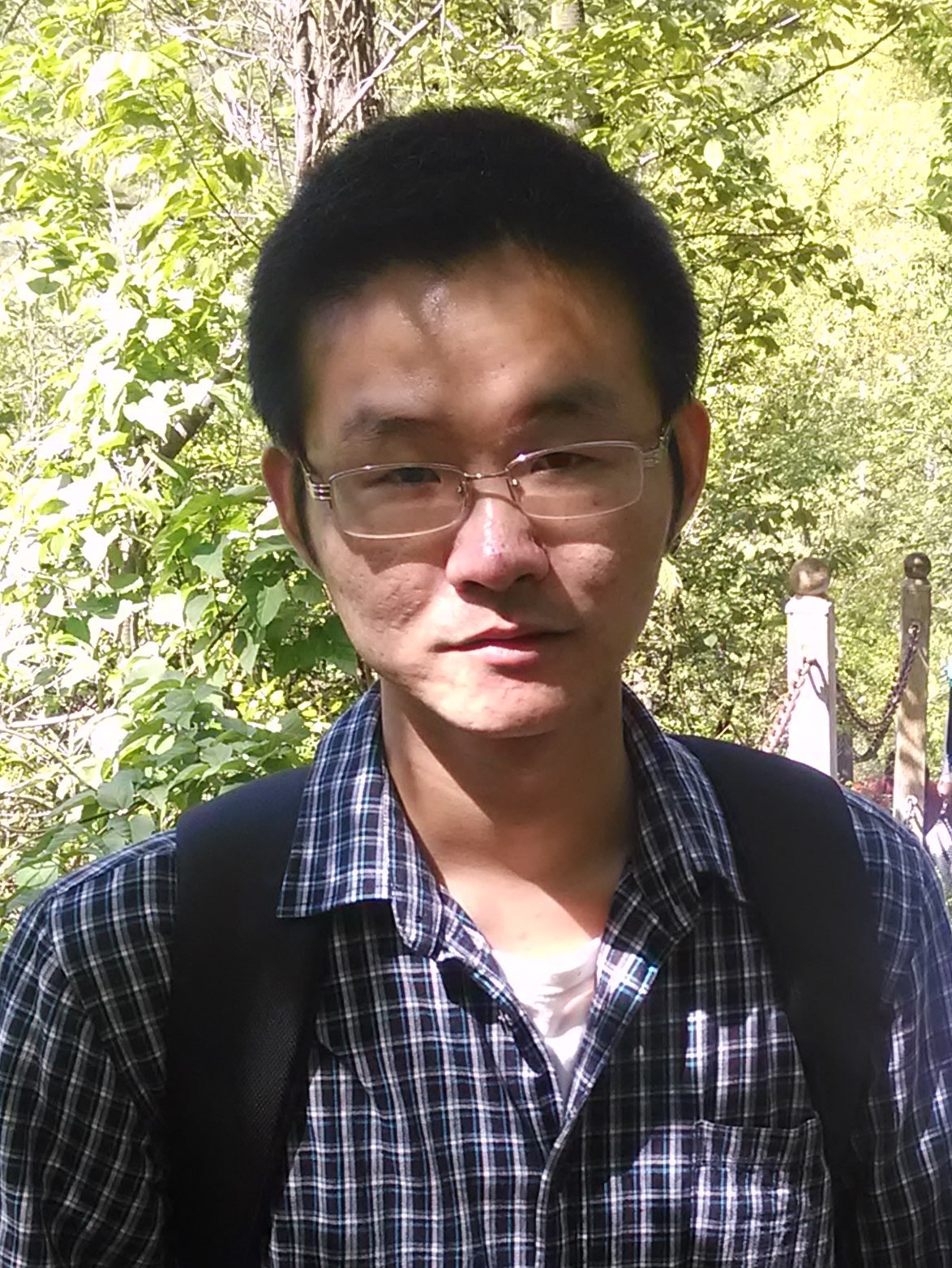}}]{Yi Wang} 
received his Ph.D. degree from the Chinese University of Hong Kong. He is currently a research scientist at Shanghai AI Laboratory. He serves as a reviewer for IJCV, Siggraph, TIP, CVPR, ICCV, ECCV, NeurIPS, ICLR, etc. His research interests include video understanding, image generation, and machine learning.
\end{IEEEbiography}

\begin{IEEEbiography}[{\includegraphics[width=1in,height=1.25in,clip,keepaspectratio]{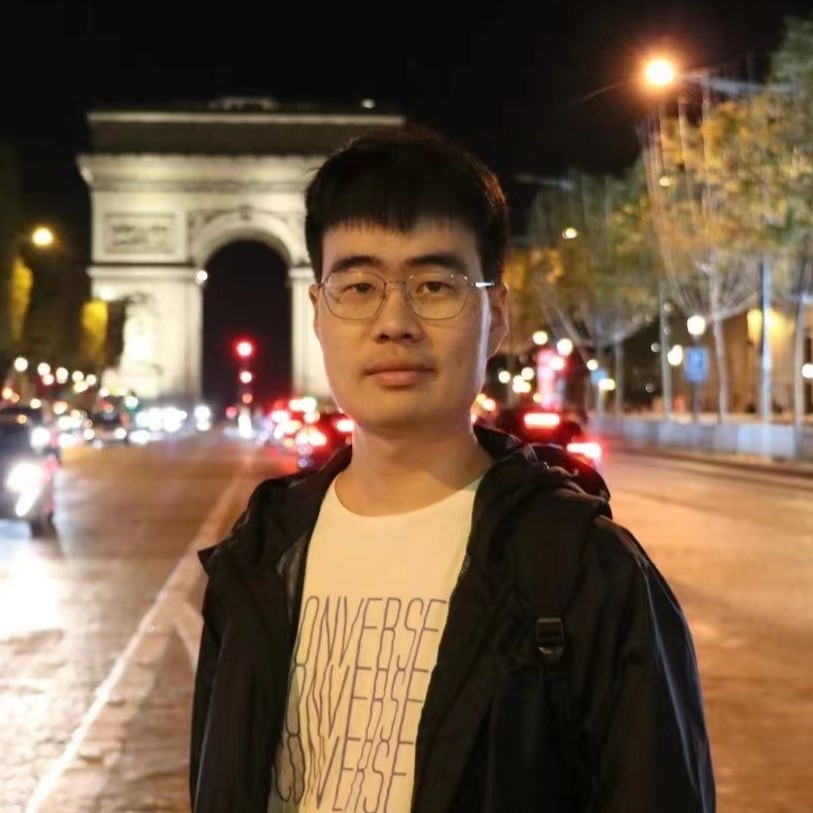}}]{Yuming Jiang} received his PhD degree from MMLab@NTU, Nanyang Technological University, supervised by Prof. Ziwei Liu and Prof. Chen Change Loy. He got his bachelor degree in computer science from Yingcai Honors College, University of Electronic Science and Technology of China (UESTC) in 2019. He received the Google PhD Fellowship in 2022. His research interests include visual generation, manipulation and restoration.
\end{IEEEbiography}

\begin{IEEEbiography}[{\includegraphics[width=1in,height=1.25in,clip,keepaspectratio]{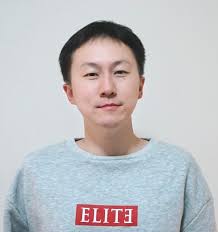}}]{Yaohui Wang} received the MS degree from Université Paris-Saclay and the PhD degree from Inria Sophia Antipolis, STARS Team, in 2021, supervised by Dr. Antitza Dantcheva and Dr. Francois Bremond. He is research scientist with Shanghai Artificial Intelligence Laboratory. His research focuses on video generation, generative modeling, and video representation learning.
\end{IEEEbiography}

\begin{IEEEbiography}[{\includegraphics[width=1in,height=1.25in,clip,keepaspectratio]{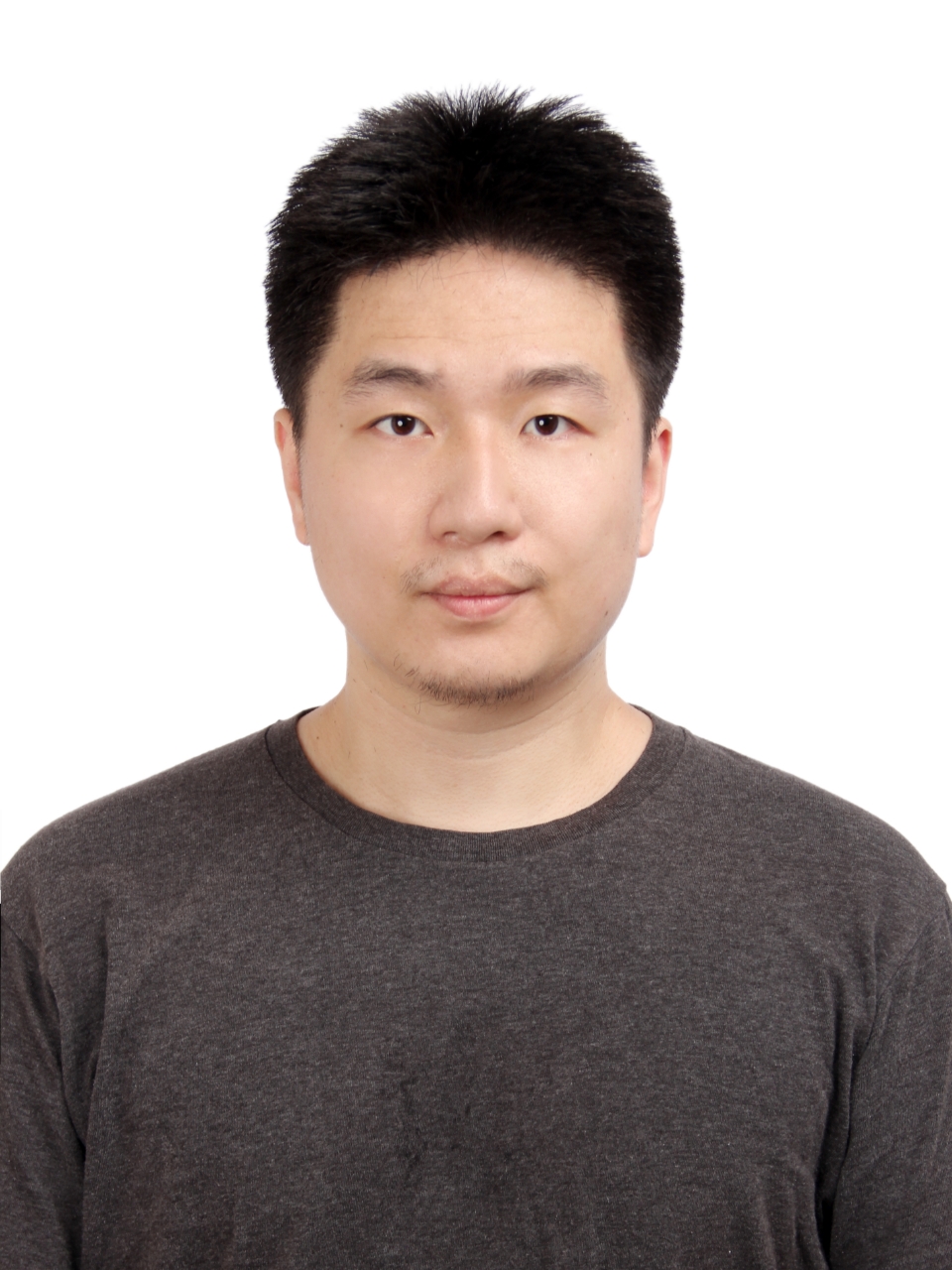}}]{Peng Gao}
is currently a research scientist at Shanghai Artificial Intelligence Laboratory. His research involves AIGCs, vision language models and large language models. He has published extensively on top-tier conferences and journals including CVPR, NerIPS, ICML and IJCV.
\end{IEEEbiography}

\begin{IEEEbiography}[{\includegraphics[width=1in,height=1.25in,clip,keepaspectratio]{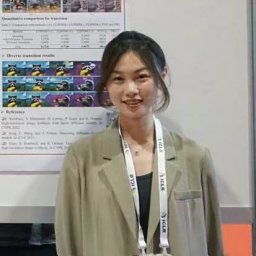}}]{Xinyuan Chen} is currently a researcher with Shanghai Artificical Intelligence Lab, China. She received dual Ph.D. degrees in Shanghai Jiao Tong University and University of Technology Sydney in 2020. Before that, she
    received the B.S. degree from Xidian University, China, in 2014.  Her research interests include video generation, generative AI, computer vision.
\end{IEEEbiography}

\begin{IEEEbiography}[{\includegraphics[width=1in,height=1.25in,clip,keepaspectratio]{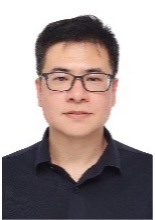}}]{Hengjie Li}
received the BS degree in computer science and technology from Harbin Institute of Technology, Harbin, in 2006, and the PhD degree in computer science and technology from Institute of Computing Technology, Chinese Academy of Sciences, Beijing, in 2013. He is a young leading scientists of Shanghai AI Lab. His research interests include parallel computing and AI system.
\end{IEEEbiography}

\begin{IEEEbiography}[{\includegraphics[width=1in,height=1.25in,clip,keepaspectratio]{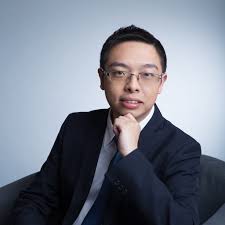}}]{Dahua Lin} received the B.Eng. degree from the University of Science and Technology of China, Hefei, China, in 2004, the M. Phil. degree from The Chinese University of Hong Kong, Hong Kong, in 2006, and the Ph.D. degree from the Massachusetts Institute of Technology, Cambridge, MA, USA, in 2012. From 2012 to 2014, he was a Research Assistant Professor with Toyota Technological Institute at Chicago, Chicago, IL, USA. He is currently an Associate Professor with the Department of Information Engineering, The Chinese University of Hong Kong (CUHK), and the Director of CUHK-SenseTime Joint Laboratory. His research interests include computer vision and machine learning.
\end{IEEEbiography}

\begin{IEEEbiography}[{\includegraphics[width=1in,height=1.25in,clip,keepaspectratio]{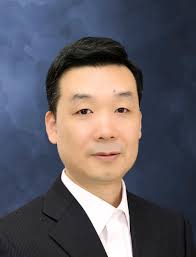}}]{Yu Qiao} (Senior Member, IEEE) is a professor and Leading Scientist at the Shanghai AI Laboratory, previously the Director of the Institute of Advanced Computing and Digital Engineering at the Shenzhen Institutes of Advanced Technology, Chinese Academy of Science. He has published more than 300 research papers with more than 79,000 citations. His team won the AAAI 2021 Best Paper, the CVPR 2023 Best Paper, and the ACL 2024 Distinguished Paper. He received the Young Scholar Award of Wang Xuan Award, the First Prize of the Guangdong Technological Invention Award, and the Jiaxi Lv Young Researcher Award from the Chinese Academy of Sciences. His research interests include deep learning on computer vision, video understanding and generation, and multimodal large models.
\end{IEEEbiography}

\begin{IEEEbiography}
[{\includegraphics[width=1in,height=1.25in,clip,keepaspectratio]{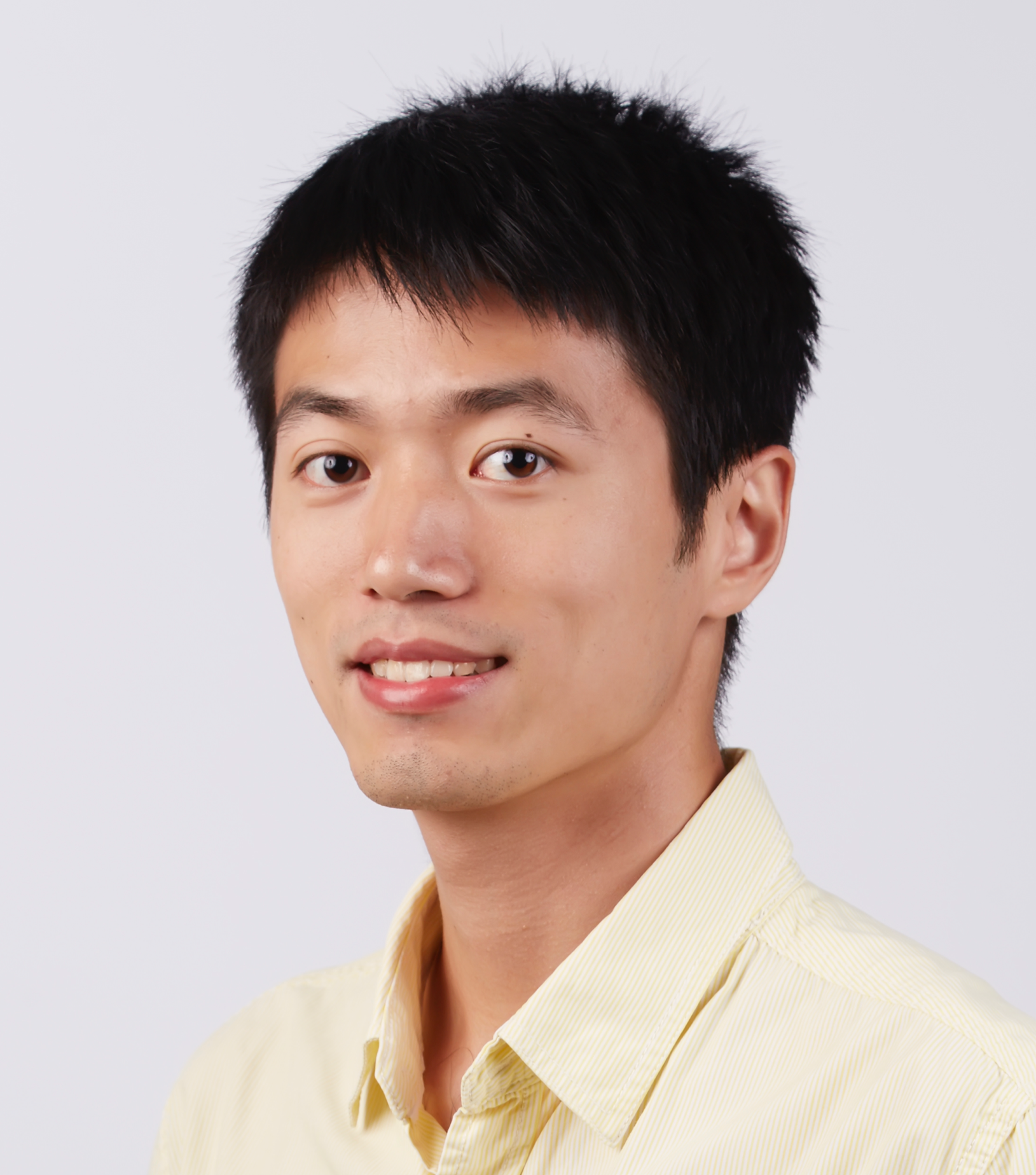}}]{Ziwei Liu} is currently a Nanyang Assistant Professor at Nanyang Technological University, Singapore. His research revolves around computer vision, machine learning and computer graphics. He has published extensively on top-tier conferences and journals in relevant fields, including CVPR, ICCV, ECCV, NeurIPS, ICLR, ICML, TPAMI, TOG and Nature - Machine Intelligence. He is the recipient of Microsoft Young Fellowship, Hong Kong PhD Fellowship, ICCV Young Researcher Award, HKSTP Best Paper Award and WAIC Yunfan Award. He serves as an Area Chair of CVPR, ICCV, NeurIPS and ICLR, as well as an Associate Editor of IJCV.
\end{IEEEbiography}
\end{document}

%% file: sec/1_introduction.tex

\section{Introduction}
\label{sec:introduction}

\begin{figure*}[t]
\centering
\includegraphics[width=1.0\textwidth]{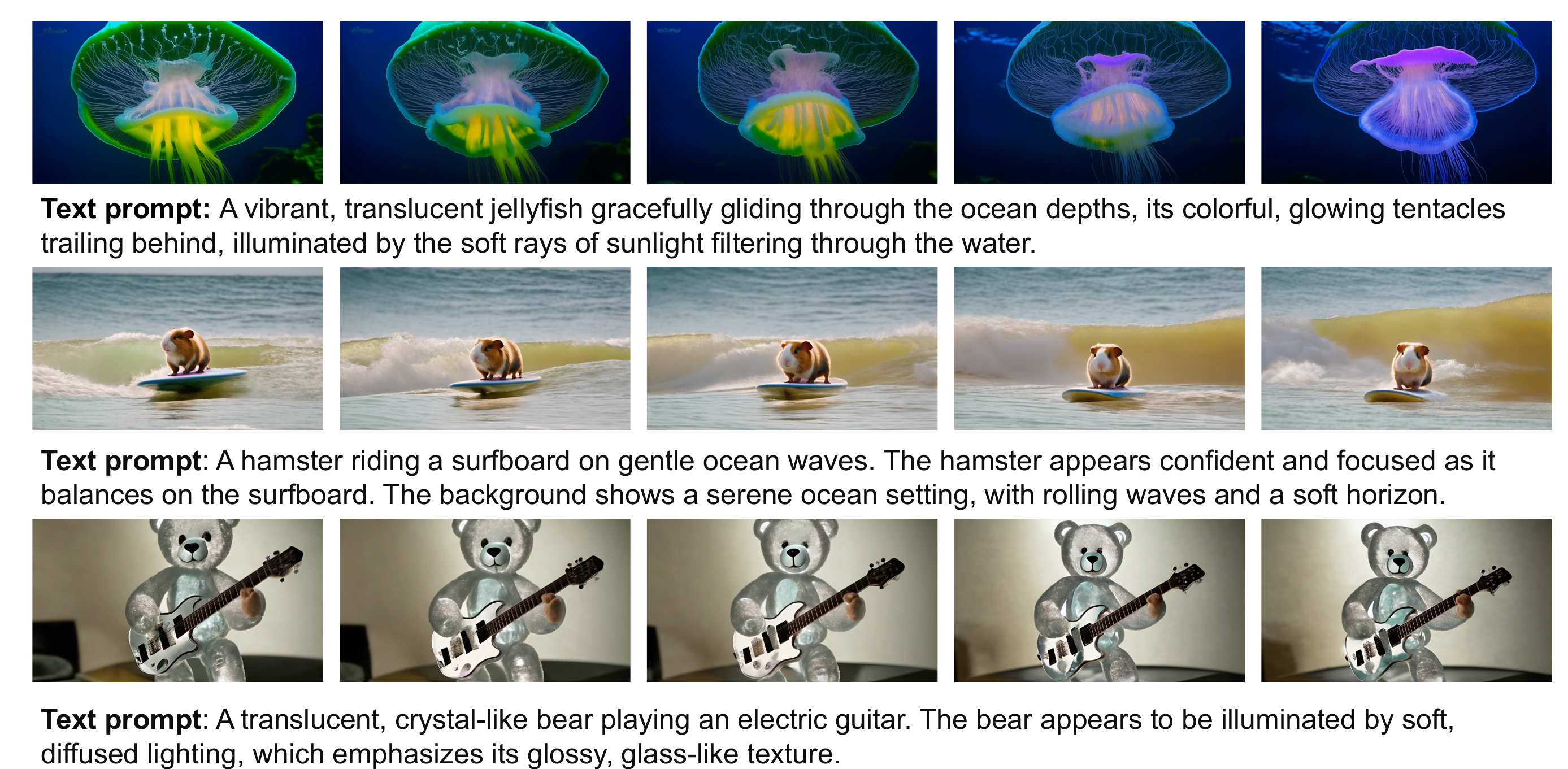}
\caption{High quality videos generated by our 2B model, demonstrating its capabilities in generating videos with high-fidelity and consistent actions.}
\label{fig:teaser}
\end{figure*}

\IEEEPARstart{R}{ecent} advancements in generative models have significantly enhanced text-to-image (T2I) synthesis, with diffusion models emerging as a leading approach due to their ability to produce high-fidelity, semantically rich images through iterative denoising processes~\cite{esser2021taming,ramesh2021zero,ding2021cogview,lin2021m6,yu2022scaling,ho2020denoising,song2020denoising,rombach2022high,peebles2023scalable,chen2023pixart,podellsdxl}. Building upon this success, researchers have extended diffusion models to text-to-video (T2V) generation, aiming to create temporally coherent and visually compelling videos from textual descriptions~\cite{singer2022make,ho2022imagen,an2023latent,zhou2022magicvideo,blattmann2023align}.

Transitioning from image to video generation introduces unique challenges. Unlike static images, videos require not only high spatial fidelity within individual frames but also seamless temporal consistency across sequences. This necessitates sophisticated modeling of dynamic interactions and smooth transitions between frames. Additionally, the multi-frame nature of video data substantially increases computational and memory demands, underscoring the need for scalable and efficient models. Limited availability of high quality annotated video datasets is another major challenge, which hampers the diversity and representativeness of training data, thereby restricting the generalization capabilities of existing models. Therefore, we introduce the \textbf{Vchitect T2V DataVerse}, a carefully curated high-quality dataset, which covers a wide variety of high quality and high fidelity videos. Developed through detailed annotation and aesthetic evaluation processes, this dataset ensures that text-video alignment is preserved across diverse and complex tasks. This dataset significantly enhances our model’s training and generalization capabilities.

Current T2V generation methods, such as Make-A-Video~\cite{singer2022make}, Imagen Video~\cite{ho2022imagen}, and MagicVideo~\cite{zhou2022magicvideo}, have adapted T2I diffusion models by incorporating temporal modules like spatio-temporal attention, directed temporal attention, or 3D convolutions. While these techniques have facilitated progress in generating short, high-resolution video clips, they face significant limitations when scaling to longer video sequences. Temporal inconsistencies, including flickering and motion discontinuities, often emerge, compromising the coherence of generated videos. Moreover, the substantial computational costs associated with training and inference limit the applicability of these methods in real-world scenarios that demand long-duration or high-resolution videos. Notably, many existing approaches focus on scaling up model size or dataset scale rather than optimizing architectures for more efficient temporal modeling.

To address these limitations, we propose \textbf{Vchitect-2.0}, an innovative parallel transformer architecture specifically designed to tackle the challenges inherent in video diffusion models. At its core, Vchitect-2.0 features a multimodal diffusion block that ensures robust alignment between text prompts and frame-wise features while maintaining temporal consistency across video sequences. By employing a hybrid parallelism framework that combines data and sequence parallelism with advanced memory optimization techniques such as recomputation and offloading, Vchitect-2.0 achieves scalability and efficiency without sacrificing quality. This framework facilitates efficient training on distributed systems, enabling the generation of long-duration, high-resolution videos while effectively addressing memory bottlenecks.


We validate the effectiveness of Vchitect-2.0 through extensive experiments. Benchmark evaluations demonstrate that our model consistently outperforms state-of-the-art approaches across key metrics, including temporal coherence, spatial fidelity, and computational efficiency. Qualitative analyses reveal that Vchitect-2.0 produces smoother transitions between frames, mitigating motion artifacts such as flickering and discontinuities, while preserving fine-grained spatial details. Furthermore, the hybrid parallelism framework enhances training scalability, enabling faster convergence and reduced memory overhead compared to existing methods. Ablation studies underscore the significance of the multimodal diffusion block and parallelism strategies in achieving these outcomes, offering insights into their roles in improving overall performance.

Through these advancements, Vchitect-2.0 establishes a new benchmark in text-to-video generation by addressing the critical limitations of existing approaches and providing a scalable, efficient framework that excels in both quality and performance. By integrating architectural innovations, a robust training pipeline, and a high-quality dataset, our work lays a solid foundation for future research in video generative models.

Our contributions are summarized as follows:
\begin{itemize}
\item Introduction of \textbf{Vchitect-2.0}, a scalable parallel transformer architecture that effectively addresses challenges in temporal consistency, scalability, and computational efficiency within video diffusion models.
\item Development of a \textbf{hybrid parallelism framework} that integrates spatial and temporal attention mechanisms with memory optimization techniques, facilitating efficient training of extended video sequences.
\item Creation of the \textbf{Vchitect T2V DataVerse}, a comprehensive dataset that ensures alignment between textual descriptions and video content, supporting a wide range of complex T2V tasks.
\item Execution of extensive experiments, including ablation studies and qualitative evaluations, demonstrating state-of-the-art performance across benchmarks and providing insights into the efficacy of our approach.
\end{itemize}

%% file: sec/2_related_work.tex
\section{Related Works}
\subsection{Text-to-video and Image-to-video Models}
Recent advancements in diffusion models~\cite{Diffusion, DiffusionModel} and large-scale text-video datasets~\cite{panda70m} have significantly propelled the field of video generation. These models generate videos either from text descriptions or image inputs. For instance, Imagen Video~\cite{Imagen} employs a cascading structure for generating high-resolution videos, while the Video Diffusion Model~\cite{videodiffusion} extends standard image generation architectures to handle video data, enabling training on both image and video datasets. Building on text-to-image (T2I) models like Stable Diffusion~\cite{LDM}, several approaches incorporate additional layers to ensure cross-frame motion consistency. Notable methods include Tune-A-Video~\cite{tuneavideo}, which utilizes a causal attention mechanism to reduce computational costs, and Align-Your-Latents~\cite{align}, which transforms T2I models into video generators by aligning independently sampled noise maps. AnimateDiff~\cite{animatediff} introduces a plug-and-play temporal module for generating videos from personalized image models~\cite{sd1.5}. Other approaches explore innovative architectures, such as integrating latent and pixel spaces~\cite{show-1} or adopting cascaded generation frameworks~\cite{lavie}. For image-to-video tasks, methods like I2VGenXL~\cite{I2Vgenxl}, VideoCrafter~\cite{videocrafter1, videocrafter2}, and others achieve state-of-the-art results in terms of pixel quality and temporal coherence.

\subsection{Sequence Parallelism}
Sequence parallelism (SP) techniques have been developed to address the significant memory demands of long-context transformers. These approaches can be categorized into head parallelism (HP) and context parallelism (CP), based on their strategy for distributed attention. HP~\cite{jacobs2023deepspeed} shreds sequences by attention heads, using synchronized all-to-all communication around the attention kernel to keep sequences local for computation. Alternatively, CP~\cite{liu2023ring,nvidia_context_parallelism} adopts blockwise attention, processing local output blocks while leveraging global KV blocks through ring-style or all-gather communication.

While HP is limited by the number of attention heads, CP suffers from cross-node communication overhead. To address these limitations, methods like USP~\cite{USP} combine inter-node ring-attention with intra-node Ulysses to enhance scalability. LoongTrain~\cite{LoongTrain} further introduces a flexible configuration for CP and HP sizes, enabling longer context lengths and improved training throughput. These advancements make SP a critical enabler for handling long video sequences.

\subsection{Memory-Efficient Training}
Since SP does not partition model parameters, various techniques have been introduced to optimize memory usage. Among them, ZeRO~\cite{ZeRO} optimization is widely employed to distribute parameters, gradients, and optimizer states across devices, reducing communication overheads. Other strategies, such as tensor parallelism (TP)\cite{megatron} and pipeline parallelism (PP)\cite{GPipe,PipeDream}, require substantial model refactoring and are more complex to integrate with SP. Recent implementations of ZeRO~\cite{pytorch_distributed,fsdp} mitigate these challenges by replicating model parameters locally within each node, balancing memory usage and communication efficiency.

Single-device memory optimization techniques complement SP to further reduce memory overhead. Mixed precision training~\cite{mixed_precision} uses lower precision (16-bit) for most computations while maintaining 32-bit precision for critical layers. Selective gradient checkpointing~\cite{Checkpoint} trades memory for computation by recomputing activations during backpropagation rather than storing them. Activation offloading~\cite{Swap} shifts activations to host memory with prefetching to minimize latency, although this is constrained by CPU-GPU bandwidth. Activation quantization~\cite{QDiffusion,ActNN} compresses activations but risks accuracy loss due to packing and unpacking overheads. Additionally, efficient kernels such as FlashAttention~\cite{flash-attention} reduce activation storage requirements, further optimizing large-scale training workflows.

\begin{figure*}[t]
\centering
\includegraphics[width=1.0\textwidth]{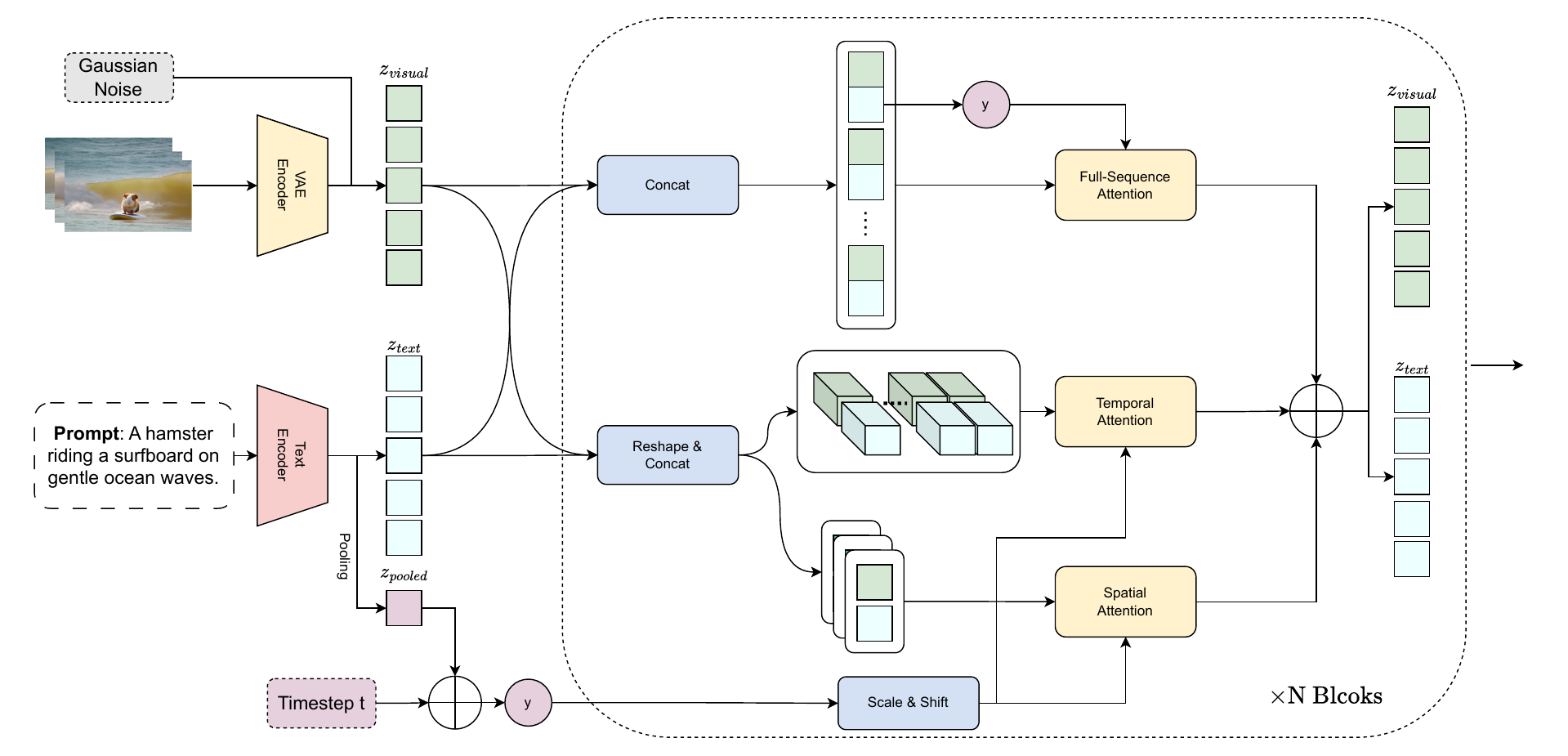}
\caption{Overview of the parallel transformer architecture used in Vchitect-2.0. The model combines text and video features using a unified framework. The \textit{Text Encoder} processes input prompts, generating textual embeddings ($z_\text{text}$), while the \textit{VAE Encoder} extracts visual embeddings ($z_\text{visual}$) from video frames. These embeddings are concatenated and processed through a series of attention mechanisms, including \textit{Full-Sequence Attention}, \textit{Temporal Attention}, and \textit{Spatial Attention}, which maintain both spatial and temporal consistency.}
\label{fig:method_T2V}
\end{figure*}

%% file: sec/3_method.tex
\section{Method}
\subsection{Preliminaries}

Denoising Diffusion Probabilistic Models (DDPMs) \cite{ho2020denoising} and Latent Diffusion Models (LDMs) \cite{rombach2022high} form the foundational frameworks for modern generative modeling. These models leverage diffusion processes for data distribution modeling and achieve efficient synthesis by operating in latent spaces.

DDPMs model the data distribution using a two-phase process: forward diffusion and reverse denoising. In the forward process, Gaussian noise is progressively added to the data $\boldsymbol{x}_0 \sim q(\boldsymbol{x}_0)$ through a Markov chain:
\begin{align}
    q(\boldsymbol{x}_{1:T}|\boldsymbol{x}_0) = \prod_{t=1}^T q(\boldsymbol{x}_t|\boldsymbol{x}_{t-1}),
\end{align}
\begin{align}
    q(\boldsymbol{x}_t|\boldsymbol{x}_{t-1}) = \mathcal{N}(\boldsymbol{x}_t;\sqrt{1-\beta_t}\boldsymbol{x}_{t-1},\beta_t \mathcal{I}),
\end{align}
where $\beta_t$ defines the noise variance schedule. In the reverse process, the noisy sample $\boldsymbol{x}_T$ is iteratively refined to reconstruct $\boldsymbol{x}_0$:
\begin{align}
    p_\theta(\boldsymbol{x}_{0:T}) = p(\boldsymbol{x}_T) \prod_{t=1}^T p_\theta(\boldsymbol{x}_{t-1}|\boldsymbol{x}_t),
\end{align}
\begin{align}
    p_\theta(\boldsymbol{x}_{t-1}|\boldsymbol{x}_t) = \mathcal{N}(\boldsymbol{x}_{t-1};\boldsymbol{\mu}_\theta(\boldsymbol{x}_t, t), \boldsymbol{\Sigma}_\theta(\boldsymbol{x}_t, t)),
\end{align}
where $\boldsymbol{\mu}_\theta$ and $\boldsymbol{\Sigma}_\theta$ are estimated by a time-conditional UNet, trained to minimize the objective:
\begin{align}
    \mathcal{L} = \mathbb{E}_{\boldsymbol{x}, \epsilon \sim \mathcal{N}(0,1), t} \left[ \Vert \epsilon - \epsilon_\theta(\boldsymbol{x}_t, t) \Vert_2^2 \right].
\end{align}

LDMs \cite{rombach2022high} extend DDPMs by operating in a learned latent space instead of the original pixel space, significantly reducing computational demands. In the first stage, an autoencoder compresses the input $\boldsymbol{x} \in \mathbb{R}^{H \times W \times 3}$ into a latent representation $\boldsymbol{z} = \mathcal{E}(\boldsymbol{x}) \in \mathbb{R}^{h \times w \times c}$ via an encoder $\mathcal{E}$, and reconstructs it using a decoder $\mathcal{D}$. The diffusion model is then trained in the latent space to denoise $\boldsymbol{z}_t$, minimizing:
\begin{align}
    \mathcal{L} = \mathbb{E}_{\boldsymbol{z}, \epsilon \sim \mathcal{N}(0,1), t} \left[ \Vert \epsilon - \epsilon_\theta(\boldsymbol{z}_t, t) \Vert_2^2 \right].
\end{align}
This latent-space operation reduces parameter count and memory consumption while maintaining performance, making LDMs efficient for generative tasks.

\subsection{Parallel Transformer Architecture}
Our model builds upon the foundation of Stable Diffusion 3~\cite{esser2403scaling}, replacing traditional self-attention and cross-attention layers with unified multimodal diffusion blocks, augmented by the insertion of learnable context latents into the sequence. This modification enhances text-frame coherence but cannot be directly extended to text-video generation due to the need for maintaining two distinct types of coherence: text-image and text-video. To address this, we introduce a parallel multimodal diffusion block, which retains the inherent text-image generation capability while achieving consistent and generalized text-to-video synthesis.

An overview of the proposed structure is depicted in Figure~\ref{fig:method_T2V}. While retaining the MM-DiT block from Stable Diffusion 3, originally designed for 2D images, our parallel transformer extends its functionality to handle both image and video data uniformly. This is achieved by incorporating a temporal attention mechanism alongside a full-sequence cross-attention module. Given a text embedding $Z_{text}=[z_{text},z_{text},…,z_{text}]$ and a visual embedding $Z_{visual}=[z_{visual},z_{visual},…,z_{visual}]$, the embeddings are interleaved in a checkerboard pattern as $[z_{text},z_{visual},…,z_{text},z_{visual}]$ for full-sequence cross-attention, which is anchored by the text embedding from the first frame.

For spatial and temporal attention, the input sequences, shaped as $[\textit{B}, \textit{F}, \textit{L}, \textit{H}, \textit{W}]$ (where $\textit{B}$ denotes batch size, $\textit{F}$ is the number of frames, $\textit{L}$ is token length per frame, and $\textit{H}$ and $\textit{W}$ represent spatial dimensions), are transformed for further processing. Spatial attention reshapes the input to $[\textit{B}\times\textit{F}, \textit{L}, \textit{H}, \textit{W}]$, while temporal attention reshapes it to $[\textit{B}\times\textit{L}, \textit{F}, \textit{H}, \textit{W}]$. The outputs of the spatial, temporal, and full-sequence cross-attention branches are then aggregated via element-wise addition, producing the final feature representation.

\section{Vchitect T2V DataVerse}
\label{sec:data}
Vchitect 2.0 builds upon previous work with longer video durations and a more rigorous data annotation process. We utilized publicly available datasets, such as WebVid10M~\cite{webvid10m}, Panda70M~\cite{panda70m}, Vimeo25M~\cite{lavie} and InternVid~\cite{internvid}, implementing filtering and re-annotation. Additionally, we incorporated 1 million internally sourced videos with more controllable quality.

\begin{figure*}[t]
\centering
\includegraphics[width=1.0\textwidth]{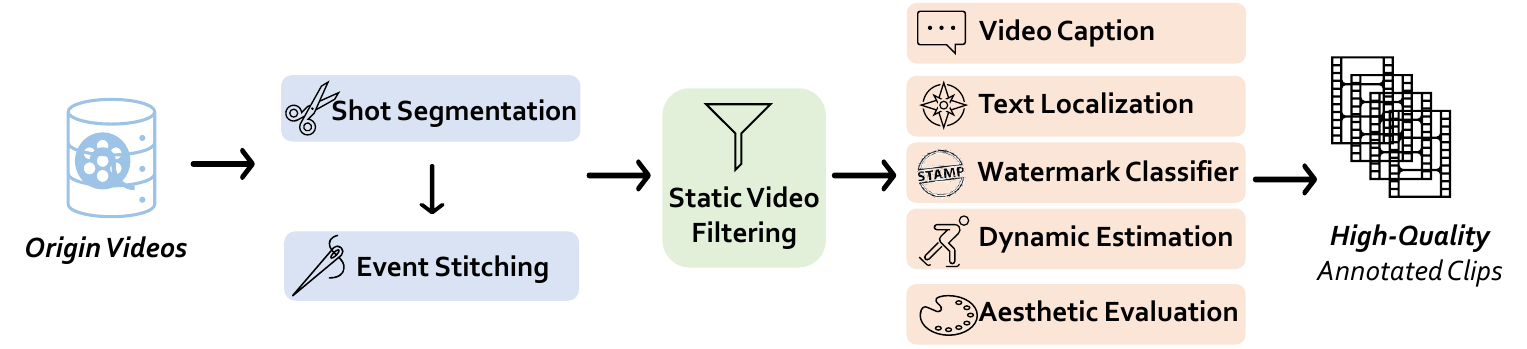}
\caption{Overview of the data processing pipeline. The pipeline begins with \textit{Origin Videos}, which are segmented into smaller \textit{Shots} using scene detection algorithms. Related shots are then merged during the \textit{Event Stitching} phase to ensure narrative coherence. \textit{Static Video Filtering} eliminates clips lacking significant motion, refining the dataset to focus on dynamic content. Subsequent processes include \textit{Video Captioning} for descriptive annotations, \textit{Text Localization} to detect text regions, a \textit{Watermark Classifier} to identify and handle watermarked content, \textit{Dynamic Estimation} to evaluate motion, and \textit{Aesthetic Evaluation} to assess visual quality.}
\label{fig:data_pipeline}
\end{figure*}

\subsection{Data Collection Pipeline}
The data processing pipeline starts with the collection of origin videos, which serve as the foundational data for the subsequent steps, as shown in Figure \ref{fig:data_pipeline}. The first stage is \textbf{Shot Segmentation}, where long videos are divided into shorter, manageable clips. Following this, \textbf{Event Stitching} links related clips to form coherent events, maintaining the narrative flow. Once the clips are organized, we apply \textbf{Static Video Filtering} to eliminate clips lacking significant motion, ensuring that video focuses on dynamic content. After filtering, we perform \textbf{Aesthetic Evaluation} to assess the visual quality of the clips, scoring them based on predetermined aesthetic criteria. Simultaneously, \textbf{Dynamic Estimation} analyzes the motion dynamics within the clips, helping to understand movement patterns and overall dynamics. In parallel, \textbf{Video Captioning} generates the corresponding paragraph pairs for text-to-video training. To ensure content integrity, the \textbf{Watermark Classifier} is used to identify whether a video contains watermarks and to fix the appropriate annotations. Additionally, \textbf{Text Localization} involves detecting the locations of any text appearing in the video, excluding watermarks, to facilitate subsequent training.  The culmination of these processes results in high-quality annotated clips.

\textbf{Shot Segmentation.}
The videos we obtained typically last several minutes or even hours. Initially, we used  \texttt{PySceneDetect}~\cite{pyscenedetect} to segment the videos into different shots based on the pixel change information. An \texttt{AdaptiveDetector} with a threshold of 21.0 was utilized to detect scene transitions.

\textbf{Event Stitching.}
Since we aim to segment videos into scenes with different semantic content, \texttt{PyScenedetect} often results in a number of false positive segmentations. Inspired by Panda-70M~\cite{panda70m}, we need to re-identify segments that were incorrectly split from the same video. Unlike the original approach, our scene identification relies on video segments, so we utilize ViCLIP~\cite{internvid}, pre-trained on video segments, instead of the initial image feature extractor. We compare the ViCLIP features of the last 8 frames and the first 8 frames of each segment; if the similarity exceeds 0.6, the video segments are considered to belong to the same scene.

\textbf{Static Video Filtering.} This process converts frames to grayscale and compares the mean pixel intensity between consecutive frames. If the difference is below a predefined threshold, the frame is considered static. By maintaining a count of static frames, we calculate their ratio to the total number of frames. Clips that exceed a specified static ratio are filtered out. We set the threshold for the mean intensity difference at 0.9 and retain video clips with less than 30\% static frames to ensure that only clips with meaningful motion are preserved. This method can be efficiently executed using only the CPU, allowing for rapid filtering of large volumes of unnecessary videos, ultimately enhancing the relevance and quality of our dataset.

\textbf{Aesthetic Evaluation.} After obtaining video clips of single events, aesthetic evaluation is a critical component of our annotation process. To ensure that the generated videos possess high artistic quality, it is essential to utilize well-lit, well-composed, and clear video segments. We employ an Aesthetics Predictor~\cite{aes}, pre-trained on 441k samples with various aesthetic annotations, to annotate the video frames. This predictor performs regression on the CLIP features to generate aesthetic scores on a scale from 1 to 10. The resulting scores will be retained as criteria for subsequent filtering.

\textbf{Dynamic Estimation.}
Excessive camera shake and minimal changes in the scene can negatively impact the training of video generation models. Although we have filtered out static videos as much as possible, there may still be overlooked low-quality videos (such as those that are originally in black and white). This module references the definition of Dynamic Degree from VBench~\cite{vbench}, utilizing RAFT~\cite{raft} to estimate the dynamic degree of the existing videos. To ensure speed, the input resolution is downsampled to 128, with the iteration set to 2.

\textbf{Video Captioning.} In training for the text-to-video task, obtaining detailed and accurate video descriptions is a crucial processing step. We utilize the trained LLaVA-Next-Video~\cite{llavanextvideo} and fine-tuned VideoChat~\cite{videochat} models to annotate the processed video clips. For both models, we use an input of 16 frames. The prompt used is: 
\begin{quote}\texttt{Please provide a detailed description of the video, focusing on the main subjects, their actions, and the background scenes.}\end{quote} Based on the VideoChat model used in Vchitect, we incorporated instruction data from VideoChat2~\cite{videochat2} and our own collected data on shot usage. The resulting captions have an average length of 200 tokens.

\textbf{Watermark Classifier.} We use a classifier pre-trained on LAION~\cite{laion_watermark} to assess the presence of watermarks in the first, middle, and last frames of the video. Features are extracted from the video frames using EfficientNet~\cite{efficientnet}, and the classifier determines the probability of the video containing watermarks.

\textbf{Text Localization.} We utilize EasyOCR~\cite{easyocr} for text detection in videos. We apply filtering criteria, such as a detection score threshold of 0.4 and a minimum text length of 4, to ensure the validity of the results. We then calculate the area of detected text regions and compute the text area ratio relative to the total frame area. If the text area ratio exceeds 2\% for a frame, it is recorded as significant. Should the number of frames containing substantial text exceed 1, the video is marked as invalid. This approach allows for accurate localization of important text while excluding irrelevant information, providing reliable data for subsequent analysis.

\subsection{Training Data Statistics}

\begin{table*}[ht]
    \centering
    \caption{Overview of the training datasets used for Vchitect-2.0. The table summarizes five datasets, including both publicly available and in-house data. Each dataset is characterized by its domain, number of videos, average and total video duration, average caption length (original and recaptioned), and resolution. Notably, the in-house dataset features high-resolution videos (up to 4K) with significantly longer and more detailed captions, contributing to the model's ability to generate high-quality video outputs.}
    \label{tab:training_data}
    \resizebox{\textwidth}{!}{ 
        \begin{tabular}{lccccccc}
            \toprule
            Dataset & Domain & \#Videos & Avg/Total Video len & Avg caption len(ori/recap) & Resolution \\ \midrule
            WebVid-10M~\cite{webvid10m} & Open       & 10M    & 18s / 52k hr    & 12.0 words & 336p        \\
            Panda-70M~\cite{panda70m}  & Open       & 70.8M  & 8.5s / 166.8k hr  & 13.2/91.2 words & 720p        \\
            InternVid-18M-aes~\cite{internvid}  & Open       & 18M  & 8.1s / 39.5 k hr  & 32.8 words & 720p \\
            Vimeo~\cite{lavie}  & Open & 21M  & 8.6s / 52.7k hr  & 9.5/ words & 720p \\
            Inhouse data & Creative/Scene/TV & 14M & 15.82 s / 61.5 k hr & 103.3 words & 1080p/4k \\  
            \bottomrule
        \end{tabular}
    }
\end{table*}

\begin{figure*}[ht]
    \centering
    \begin{minipage}{0.55\textwidth}  
        \centering
        \includegraphics[width=\linewidth]{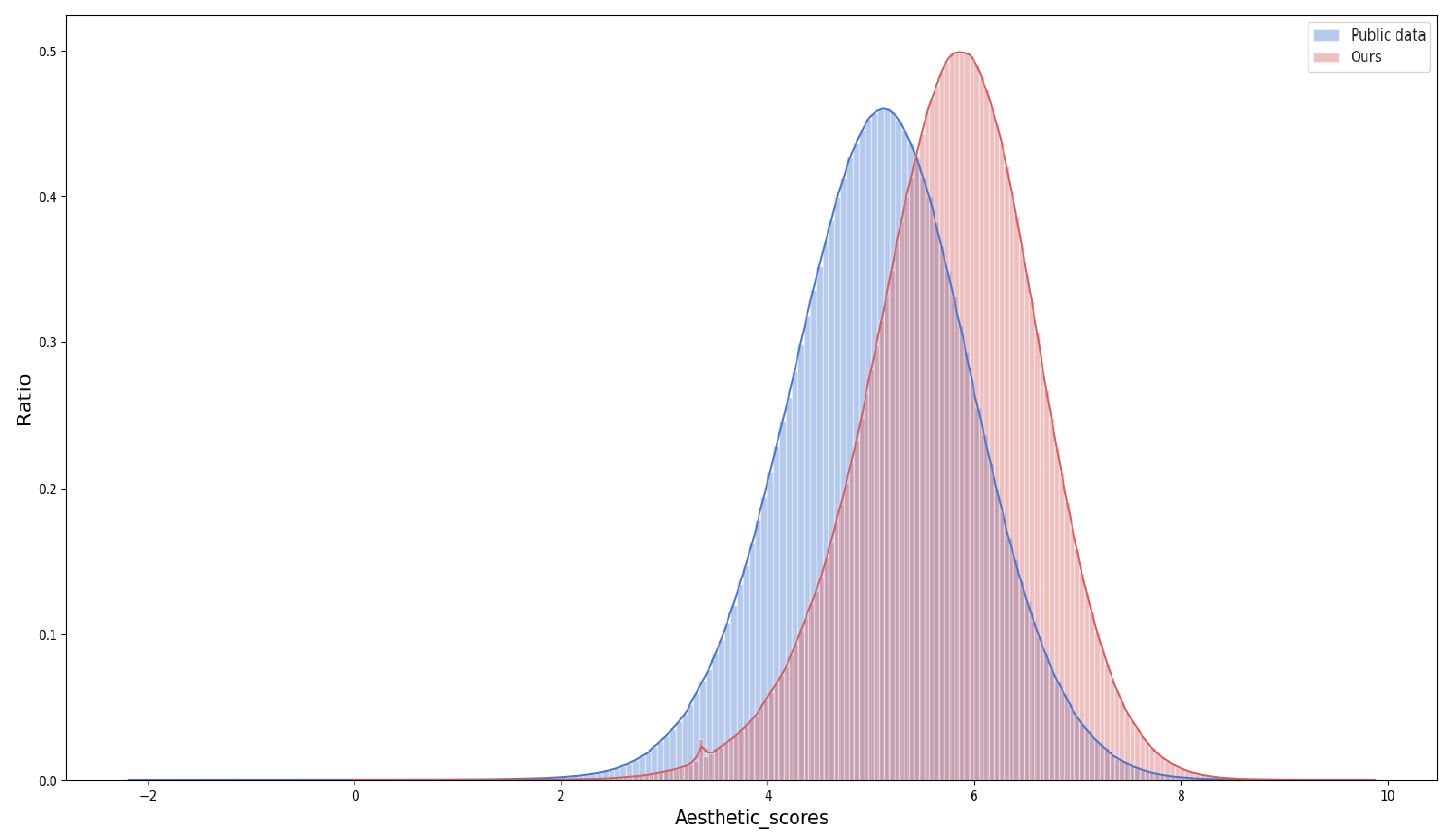}
        \caption{Comparison of the in-house data with the aesthetic distribution of public datasets.}
        \label{fig:aes}
    \end{minipage}
    \hfill
    \begin{minipage}{0.30\textwidth}  
        \centering
        \includegraphics[width=\linewidth]{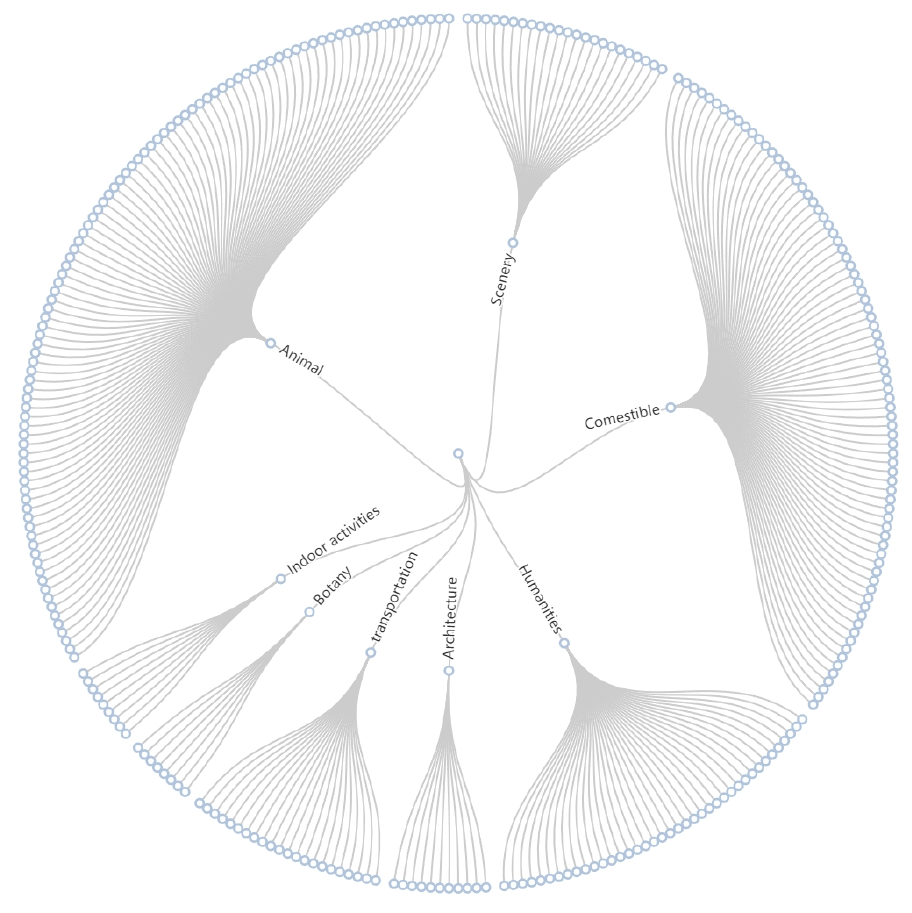} 
        \caption{Diverse categories of in-house data.}
        \label{fig:classification}
    \end{minipage}
    \label{fig:comparison}
\end{figure*}



Table~\ref{tab:training_data} shows the status of the data we prepared. Compared to Vchitect, where over half of the videos had clips shorter than 2 seconds, Vchitect-XL significantly increased the training duration of the videos. Additionally, we recaptioned the existing data, resulting in an average caption length of 100 tokens. While maintaining category diversity, we greatly improved the quality of the training data. The classification information for the in-house data is shown in Figure~\ref{fig:classification}, with the maximum resolution reaching 4K. As illustrated in Figure~\ref{fig:aes}, our data's aesthetic scores show substantial improvement over publicly available models, with nearly half of the videos scoring above 6, compared to only 16.89\% in Vchitect. We visualize the annotated data in Figure~\ ref {fig: caption}. Compared with Vchitect, the current data annotation is of higher quality, and the text description can better capture the changes of the video picture. The original annotation can only refer to the picture elements in the video, but in the Vchitect-XL annotation, in addition to capturing the camera movement, it can also describe the picture before and after the picture change in detail and capture it dynamically.

\begin{figure*}[t]
\centering
\includegraphics[width=1.0\textwidth]{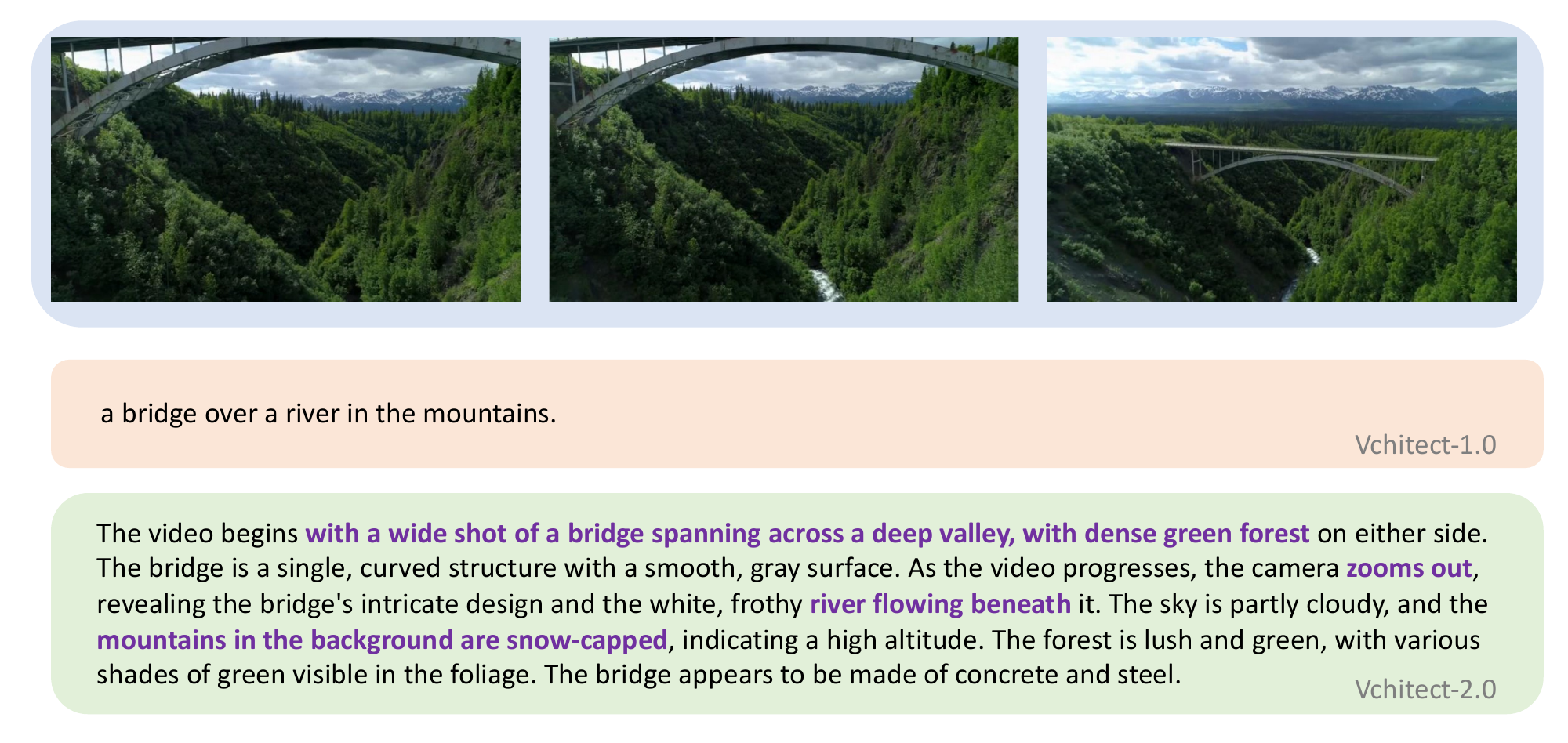}
\caption{Compared to the previous captions, the captions in Vchitect-2.0 are more detailed and can capture the changes before and after the picture and the camera movement of the video.}
\label{fig:caption}
\end{figure*}

\section{Sequence Parallelized Video Training}
\label{sec:seq}
The substantial number of vision tokens of long high-quality inputs in video training result in excessive memory footprint of forward activations. However, existing frameworks for memory reduction either face efficiency or scalability problems, or target certain types of foundation models like LLMs and are not suitable for our video model (see Sec.~\ref{subsec:training_feature} for detailed analysis). In this section, we analysis the benefits and limitations of different techniques and propose a training system that combines various memory-efficient training strategies to allow long-context video training on large clusters with gratifying training throughput. Besides optimizations targeting our specific model structure, we also integrate useful tools that might facilitate the training of different types of diffusion-based models in our open-source training framework at \href{https://github.com/Vchitect/LiteGen}{https://github.com/Vchitect/LiteGen}.

\subsection{Features of Video Training}\label{subsec:training_feature}
This section analyses the features of our video model from the training perspective and marks the limitations of existing frameworks when applied to it.

\noindent\textbf{Context length.} High-quality video data are often encoded and patchified into extremely long visual sequences due to their 3D nature, necessitating inter-node SP without other optimizations. However, since inter-node bandwidth is usually over 20 times lower than intra-node connection, either P2P~\cite{liu2023ring} or all-to-all style~\cite{Ulysses} SP techniques suffer from this slow and unstable communication when performed across training nodes. The former is bottlenecked by the slowest cross-node edge, and the latter includes synchronized collective communication which cannot be overlapped. Moreover, the scalability of all-to-all methods is restricted by the number of attention heads.

\noindent\textbf{Sparse attention.} Recent researches in LLM training propose several methods to tackle the efficiency and scalability issue of inter-node SP to facilitate multi-node training with long text sequences of over 1M tokens. However, methods like 2D attention and double RingAttention~\cite{USP,LoongTrain} are only effective with a long enough attention context to hide inter-node communication into intra-node attention computation. Unfortunately, many video diffusion models, including ours, adopt the spatial-temporal sparse attention approach for video synthesis, and neither the spatial nor the temporal context is long enough to adopt these techniques. Methods like DSP~\cite{DSP} focus on improving the SP performance of sparse video models by reducing all-to-all frequency but does not eliminate inter-node communication, and they do not apply to our model due to our parallel attention structure.

\noindent\textbf{3D and multi-modal nature.} The 3D nature and multi-modality of our sequence enlarge the design space of the parallelization strategy. Activations can be sharded along one or more dimension among the batch, spatial, temporal or channel axis. Different sharding strategy causes different needs of resharding, and can further choose multiple resharding axis. For example, SP can be performed along the spatial or temporal sequence, or even both. Attention along the sharded dimension requires extra communication to collect the entire sequence (or the entire KV block), while attention along the other axis can be performed locally in a way similar to data parallelism. Communication for attention also provides different choices. Take spatial sharding for example, all-to-all style SP can choose to switch to head-parallel or temporal-parallel for local attention. Once a certain sharding strategy is decided, device placement is the next choice. When the sequence length is long enough to necessitate inter-node sharding, putting different dimensions across nodes might affect training throughput in different ways. Moreover, while our text-to-video model performs attention across the concatenated sequence of visual and text tokens, other operators are performed on the two sequences separately, using different sets of parameters. Simplistic parallelism implementation might lead to load imbalance and other problems.

\noindent\textbf{Model parameters.} Although The memory footprint of model parameters and relevant gradients, optimizer states and ema states is relatively small compared with activation memory, techniques to eliminate this part of memory usage like ZeRO~\cite{ZeRO} is still worth adopting as they bring less overhead than methods to shrink activations thanks to the efficient implementation like FSDP~\cite{fsdp}.

Based on the above observations, we decide to take a simple strategy that limits SP groups within a training node by utilizing other memory saving techniques, and apply data parallelism across nodes. This strategy helps us achieve satisfying video training throughput without sophisticated configuration and code refactoring.

\subsection{Memory Efficient Video Training}\label{sec:training}
To reduce the activation memory footprint without inter-node SP, we experimented with several single-chip memory saving strategies and decided on a combined use of activation offloading and recomputation. Quantization is out of our consideration since the packing and unpacking overhead of existing systems exceed that of recomputation in our experiments, and it brings about the additional problem of model accuracy. Activation offloading is a promising technique that theoretically brings negligible overhead, and we implemented an efficient offloading framework that maximizes the overlap between memory swapping and computation with the help of cuda stream. However, our profiling results show that the bandwidth between CPU and GPU only allows a small proportion(around 1/5) of activation transfer to overlap with computation and is not enough to limit SP group within a single node. As such, we still need to adopt recomputation. Nevertheless, we are still able to utilize our offloading strategy as a substitution of recomputation in selected layers to partly reduce its overhead.

We also made specialized design to our intra-node SP workflow to fit the 3D and multi-modal sequence of our model. First, we choose to slice the spatial sequence in the Transformer backbone since it is typically longer than the temporal sequence, thus spatial slicing has less effect on computation kernel efficiency. As for the spatial attention kernel, we choose to switch to head parallelism, as intra-node SP does not meet the limitation of the head number, as switching to temporal slicing might require extra processing like padding and unpadding as video samples vary in frame numbers. This strategy also applies to image data. Text tokens are similarly sharded across all devices in an SP group, since they are concatenated to the spatial sequence for attention. Separate sharding of text and visual tokens ensures load balance in the Transformer backbone and text encoders. 

\begin{figure}[t]
\centering
\includegraphics[width=0.5\textwidth]{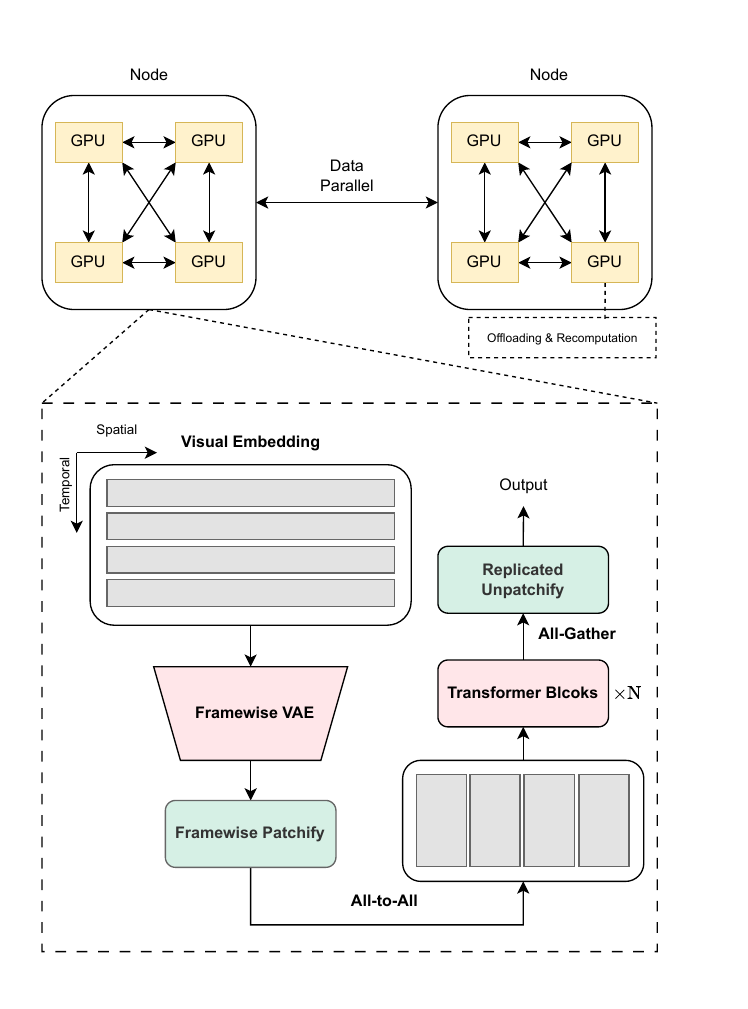}
\caption{Overview of our training workflow. The architecture integrates ZeRO-SP for memory optimization, distributed across multiple GPUs and nodes. The pipeline begins with the visual embedding, which is processed by a \textit{Framewise VAE}. The representation is passed through an all-to-all communication step and fed into $N$ Transformer Blocks to compute the features in parallel. The replicated unpatchify operations ensure that outputs are properly reassembled after the final all-gather step. The system utilizes offloading and recomputation strategies to manage memory overhead in each GPU.}
\label{fig:SP-workflow}
\end{figure}

VAE inference, as another stage in diffusion training, takes up non-negligible time in the training procedure. Moreover, although VAE memory does not affect transformer training, the excessive memory usage alone might cause OOM problems. This is because the input to VAE, the original video, is dozens times larger than the latent input into the transformer backbone, thus a few residuals quickly run out GPU memory. Consequently, it is necessary to parallelize VAE inference. We follow an existing technique to view the input video as a batch of frames and utilize frame-wise DP-VAE for inference, and slice the scattered batch of frames into smaller mini-batches and iteratively encodes a mini-batch at a time if the scattered batch is still too large. We further extend this frame-wise DP parallelism to latent pachification and position embedding calculation, and add an all-to-all resharding before entering the transformer blocks as a substitution for an all-gather after VAE encoding, which improves parallelism and reduces communication. Figure.\ref{fig:SP-workflow} shows the entire SP workflow of an iteration in our video training.

In conclusion, our training system combines intra-node SP and inter-node DP for parallelism, and integrates offloading, recomputation and FSDP to further mitigate memory footprint while minimizing consequent overheads.

\subsection{Video Training Framework}
Although our training strategies are specialized for our model and training cluster, we discovered several features of diffusion-based video models and useful gadgets to facilitate their training. Our open-source framework, LiteGen, allows flexible combination of parallelism strategies and training techniques with simple configurations and smallest code refactoring. Our framework focuses on memory optimization and parallelized training, and applies to different kinds of diffusion models, including Unet-based ones and those with multiple encoders.

%% file: sec/4_experiments.tex
\section{Experiments}
To comprehensively evaluate the performance of Vchitect-2.0, we conducted a series of experiments focusing on its ability to generate high-quality videos from textual descriptions. Our evaluation encompasses both automated and human assessments, providing a comprehensive analysis of the model's capabilities.
\begin{table*}[t]
  \centering
    \caption{Quantitative evaluation on VBench~\cite{vbench, huang2024vbenchplusplus} comparing various models across different metrics. The \textit{Total Score} represents the overall performance, while individual metrics such as \textit{Dynamic Degree}, \textit{Overall Consistency}, \textit{Aesthetic Quality}, and \textit{Imaging Quality} assess specific aspects of video generation quality. \textit{Human Action} evaluates human-centric video content, and \textit{Spatial Relationship} measures the model's understanding of spatial dynamics. Our proposed model, Vchitect-2.0-2B, demonstrates competitive results, particularly excelling in \textit{Overall Consistency}, \textit{Aesthetic Quality}, and \textit{Imaging Quality}, while the enhanced version (using VEnhancer~\cite{he2024venhancer}), Vchitect-2.0-2B [E], achieves a \textit{Total Score} comparable to the leading models.}
  \resizebox{1.0\textwidth}{!}{
  \begin{tabular}{@{}c|c|c|c|c|c|c|c}
    \toprule[1.2pt]
     Model & Total Score & Dynamic Degree & Overall Consistency & Aethetic Quality & Imaging Quality & Human Action &Spatial Relationship \\
    \midrule
    \textbf{Vchitect-2.0-2B [E]} & \textbf{82.24\%} & \underline{63.89\%} & 27.57\% & 60.41\% & 65.35\% & 97.20\% & 57.55\%\\ 
    Kling~\cite{klingai2024} & \underline{81.85\%} & 46.94\% & 26.42\% & 61.21\% & \textbf{65.62\%} & 93.40\% & \textbf{73.03\%}\\
    CogvideoX-5B~\cite{cogvideox} & 81.61\% & \textbf{70.97\%} & \underline{27.59\%} & \textbf{61.98\%} & 62.90\% & \textbf{99.40\%} & 66.35\%\\
    \textbf{Vchitect-2.0-2B}  & 81.57\% & 58.33\% & \textbf{28.01\%} & \underline{61.47\%} & \underline{65.60\%} & 97.00\% & 54.64\%  \\
    CogvideoX-2B~\cite{cogvideox} & 80.91\% & 59.86\% & 26.66\% & 60.82\% & 61.68\% & \underline{98.00\%} & \underline{69.90\%}\\
    OpenSora-v1-2~\cite{opensora} & 79.23\% & 47.22\% & 27.07\% & 56.18\% & 60.94\% & 85.80\% & 67.51\%\\
    OpenSoraPlan-v1-1~\cite{opensoraplan} & 78.00\% & 47.72\% & 26.52\% & 56.85\% & 62.28\% & 86.80\% & 53.11\%\\
    \bottomrule[1.2pt]
  \end{tabular}}
  \label{tab:vbench}
\end{table*}

\begin{table*}[htbp]
\centering
\caption{\textbf{Human Evaluation}. We show the human evaluation comparison of Vchitect-2.0-2B against other models across three key metrics: video-text alignment, frame-wise quality, and temporal quality. The table reports the number of wins against the total number of tests, and the win ratio, indicating the frequency with which our model outperforms its counterparts on each metric. The final row summarizes the total number of wins and the average win ratio of our model relative to the other four models.}
\resizebox{1.0\textwidth}{!}{%
\begin{tabular}{@{}lcccccc@{}}
\toprule
\multirow{2}{*}{\textbf{Comparison Model}}  & \multicolumn{2}{c}{\textbf{Video-Text Alignment}} & \multicolumn{2}{c}{\textbf{Frame-wise Quality}} & \multicolumn{2}{c}{\textbf{Temporal Quality}} \\
\cmidrule(lr){2-3} \cmidrule(lr){4-5} \cmidrule(lr){6-7}
                          & Wins / \#Test & Win Ratio (\%) & Wins / \#Test & Win Ratio (\%) & Wins / \#Test & Win Ratio (\%) \\ \midrule
\textbf{CogVideoX-2B}~\cite{cogvideox}     & 41 / 76   & 53.95         & 39 / 76   & 51.32         & 35 / 76   & 46.05         \\
\textbf{OpenSora-v1.2}~\cite{opensora}      & 54 / 74   & 72.97         & 50 / 74   & 67.57         & 44 / 74   & 59.46         \\
\textbf{Mira}~\cite{mira}             & 68 / 75   & 90.67         & 66 / 75   & 88.00         & 61 / 75   & 81.33         \\
\textbf{OpenSoraPlan-v1.1}~\cite{opensoraplan}  & 63 / 78   & 80.77         & 62 / 78   & 79.49         & 60 / 78   & 76.92         \\ \midrule
\textbf{Overall (Vchitect-2.0-2B)} & 226 / 303  & 74.59         & 217 / 303  & 71.62         & 200 / 303  & 66.01         \\ 

\bottomrule
\end{tabular}
}
\label{tab:human_evaluation}
\end{table*}
\subsection{Implementation Details}
Following the data strategy outlined in Section~\ref{sec:data} and the sequence training strategy detailed in Section~\ref{sec:seq}, we trained our 2B model, initialized from SD3-medium, through a five-stage process. The training began with pretraining on WebVid-10M, where zero-initialized cross-attention and temporal attention layers were employed. During this stage, both resolution and video lengths were gradually increased, starting at 240p with a single frame and scaling up to 336p with 24 frames. In the second stage, the model was fine-tuned on the Panda-70M dataset, with the video length progressively extended to 40 frames to improve its ability to handle longer sequences. The third stage focused on further extending video length and resolution, training the model with videos lasting up to 10 seconds at 240p, while shorter videos were adjusted to higher resolutions to enhance visual quality. In the fourth stage, we fine-tuned the model on InternVid-18M-aes and Vimeo datasets to support the generation of videos at resolutions up to 720p. Finally, the model was refined using high-quality internal data, concentrating on generating 5-second videos consistently at 720p to optimize the balance between temporal coherence and visual fidelity. This structured, multi-stage training approach allowed the model to scale effectively while maintaining high-quality video generation across diverse resolutions and lengths.

\subsection{Automated Evaluation}
To assess the performance of our model, we utilize all the metrics provided in VBench~\cite{vbench, huang2024vbenchplusplus}. As presented in Table~\ref{tab:vbench}, we report the \textit{Total Score} along with several representative metrics. Specifically, the \textit{Dynamic Degree} evaluates whether the videos exhibit significant motion, while \textit{Aesthetic Quality} and \textit{Imaging Quality} measure the artistic appeal and perceived visual fidelity from a human perspective. Additionally, \textit{Overall Consistency} assesses the smoothness of generated videos, and \textit{Human Action} evaluates the quality of human-centric video content.

Compared to the baseline model CogvideoX-2B, our model, Vchitect-2.0-2B, demonstrates superior performance with a higher \textit{Total Score}. Notably, our model achieves improvements of \textbf{1.35\%}, \textbf{0.65\%}, and \textbf{3.92\%} in the \textit{Overall Consistency}, \textit{Aesthetic Quality}, and \textit{Imaging Quality} metrics, respectively. These results underscore the advantages of employing a robust base model like SD3. However, CogVideoX-2B surpasses our model in the \textit{Human Action} and \textit{Spatial Relationship} metrics, highlighting current limitations in generating human-centric videos and understanding spatial dynamics.

Furthermore, we present results from an enhanced version of our model, Vchitect-2.0-2B [E], which incorporates post-processing with VEhancer~\cite{he2024venhancer}. This enhanced version achieves a \textit{Total Score} exceeding that of the commercial model Kling by \textbf{0.39\%}. These findings demonstrate the effectiveness of our approach and the potential for further refinement through post-processing techniques.
\begin{table*}[t]
  \centering
    \caption{Iteration time of memory-efficient training strategies on two ZeRO DP devices. The \textit{Baseline} approach fails for larger input shapes (marked as \textit{OOM}, out-of-memory), while \textit{Recomputation} significantly reduces memory usage at the cost of increased iteration time. \textit{Offload} is efficient for smaller inputs but leads to performance degradation or system breakdown for larger shapes. The \textit{Combination} strategy balances memory efficiency and iteration time, providing stable and scalable performance even for complex input shapes}
  \resizebox{0.8\textwidth}{!}{
  \begin{tabular}{@{}c|c|c|c|c|c}
    \toprule
     Input Shape & Selective Ratio & Baseline & Recomputation & Offload & Combination \\
    \midrule
    32×256×256 & 0.2 & 15.7 & 14.2 & 13.6 & 13.6\\
    32×512×384 & 0.5 & OOM & 26.6 & 58.4 & 25.7\\ 
    32×1024×768 & 1 & OOM & 67.5 & Break Down & 65.3\\
    128×256×256 & 0.8 & OOM & 31.2 & 130.8 & 30.3\\
    \bottomrule
  \end{tabular}}
  \label{tab:offload}
\end{table*}

\subsection{Human Evaluation}


In addition to automated evaluation, we conduct human evaluation to further assess model performance. Human evaluators are presented with pairs of videos generated by different models using the same text prompt. They are asked to indicate their preference between the two videos based on three key aspects, evaluated separately, following the dimension design in VBench~\cite{huang2024vbenchplusplus}: video-text alignment, temporal quality, and frame-wise quality. The evaluation include five models: our model, Mira~\cite{mira}, OpenSora-v1.2~\cite{opensora}, CogVideoX-2B~\cite{cogvideox}, and OpenSoraPlan-v1.1~\cite{opensoraplan}. For each aspect, a total of 303 comparisons are conducted to evaluate our model against the other four models. As shown in Tab.~\ref{tab:human_evaluation}, our model achieves an average win ratio of over 60\% relative to the other models across all three metrics. These results from the human evaluation demonstrate a clear preference for our model.

\subsection{Ablation Study}
\begin{figure}[t]
    \centering
    \includegraphics[width=0.49\textwidth]{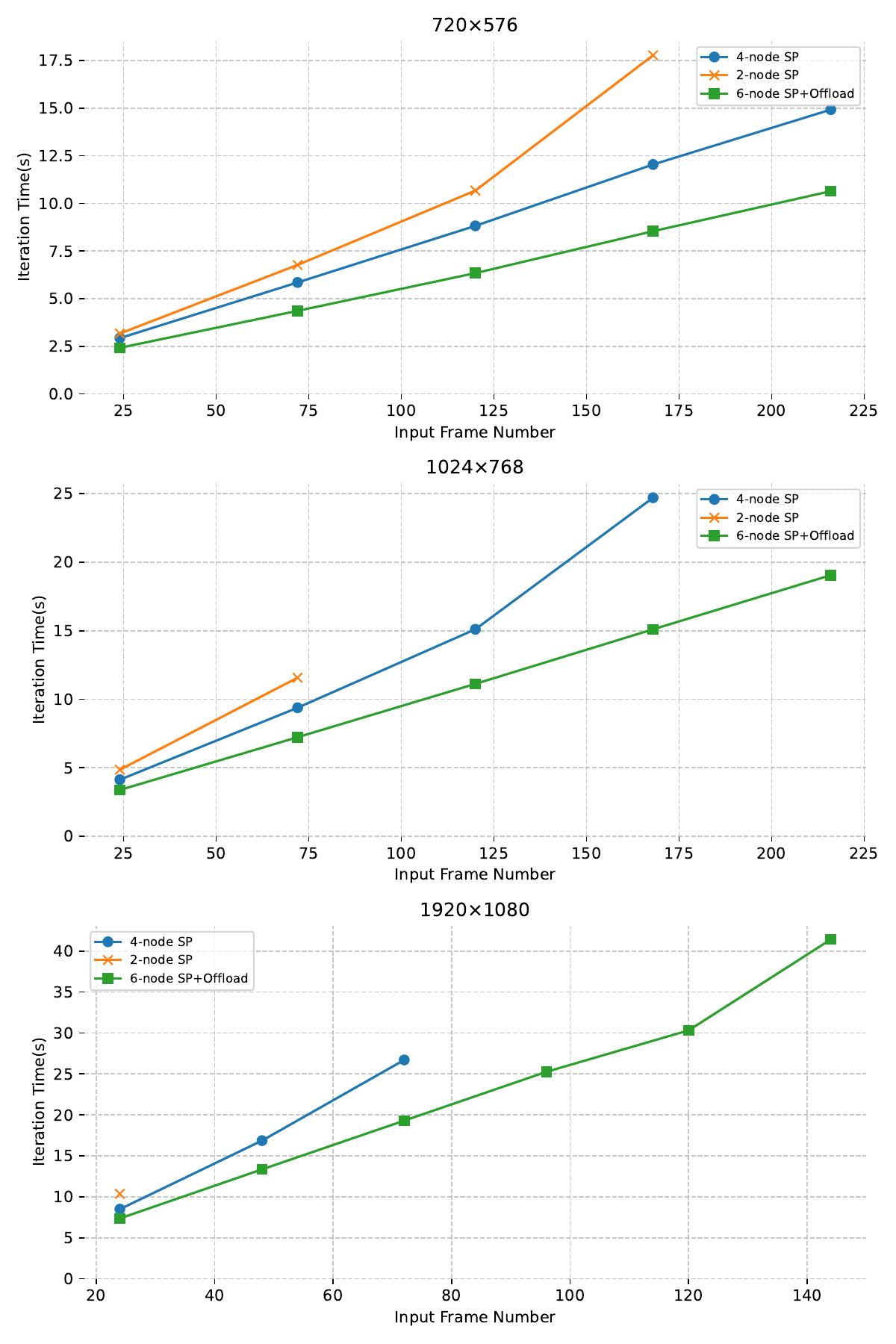}
    \caption{Iteration time of different training strategies(intra-node SP with recomputation and offload, 2-node SP and 4-node SP) according to input frame number on 4 nodes of 6 devices, with different input resolution (720×576, 1024×768 and 1920×1080 on the upper, medium and lower graph, respectively).}
    \label{fig:throughput-evaluate}
\end{figure}

\textbf{Training Throughput.} We performed our model training on our NV cluster, with training nodes of 8 A100 (80GB) GPUs, connected via intra-node NVLink and inter-node Infiniband. We evaluated the training throughput our training strategy with hybrid ZeRO-SP and DP, and extra memory elimination techniques(See Sec.\ref{sec:training}). We compared it with cross-node SP (over 2 and 4 nodes respectively) without recomputation or offload, but with intra-node ZeRO optimization to avoid the influence of parameter memory. We set a global batch size fixed to 24 input videos and gradient accumulation adopted. Our experiment is conducted on 4 training nodes, each with only 6 devices used to avoid padding on attention heads(24 in our model). Results are shown in Fig.\ref{fig:throughput-evaluate}. Our strategy not only outperforms intra-node Ulysses in training throughput over all input shapes, but scales to a much longer sequence length up to 1.16M(input shape of 144×1920×1080). In contrast, cross-node Ulysses only supports 0.6M sequence length even on the upper limit of its parallel size, equal to the number of attention heads. Moreover, while the performance of cross-node SP degrades significantly with longer input sequence, our method provides near-linear scalability related to sequence length.\\[1ex]
\begin{figure}[t]
    \centering
    \includegraphics[width=0.5\textwidth]{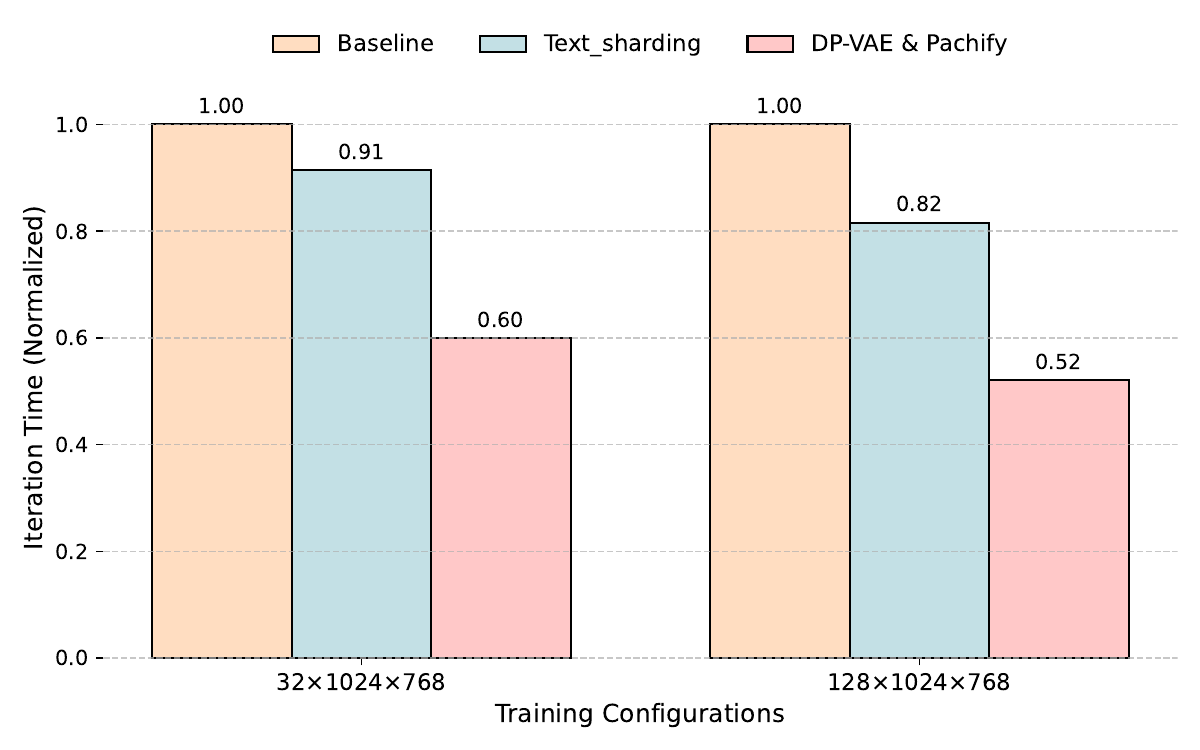}
    \caption{Iteration time of training on a training node of 8 devices, with different optimizations added progressively, regularized by baseline.}
    \label{fig:sp-evaluate}
\end{figure}
\noindent\textbf{Single-device memory efficient training techniques.} We experimented recomputation, activation offloading and their combination with different input video lengths and resolutions. We conducted our experiments on 2 cards with ZeRO-3 DP to help scale the input sequence. Results are shown in Table. For each input shape we tune the optimal selective ratio for pure recomputation and offloading. Interestingly, we find that the optimal ratio usually keeps maximum memory usage at a low level around 40-60GB. Although a higher memory footprint does not trigger OOM problems, it induces frequent memory management operations which exceed the overhead of handcrafted activation memory elimination. For combined use of the two techniques, offload ratio is fixed at 0.2, a ratio that fully utilizes the bandwidth between GPU and host memory according to our profiling results. Results in Tab.\ref{tab:offload} show that activation offload performs well under the threshold of 0.2 selective ratio (32×256×256 input), but the performance drops significantly when more activations are offloaded, which is consistent with profiling. A combination of offloading and recomputation for long input sequences reduces the overhead of simple recomputation by about 3\%. \\[1ex]
\begin{table*}[t]
  \centering
  \caption{Ablation study on model architecture. This table compares the performance of our full Vchitect-2.0 model with the baseline (without the full-sequence cross-attention module) and other competing models. The metrics include \textit{Total Score}, \textit{Dynamic Degree}, \textit{Overall Consistency}, \textit{Aesthetic Quality}, \textit{Imaging Quality}, \textit{Human Action}, and \textit{Spatial Relationship}. Our model demonstrates superior performance across most metrics, showcasing the effectiveness of the full-sequence cross-attention module.}
  \label{tab:vbench_ablation}
  \resizebox{1.0\textwidth}{!}{
  \begin{tabular}{@{}c|c|c|c|c|c|c|c@{}}
    \toprule
     Model & Total Score & Dynamic Degree & Overall Consistency & Aesthetic Quality & Imaging Quality & Human Action & Spatial Relationship \\
    \midrule
    \textbf{Ours} & \textbf{79.33\%} & \underline{43.98\%} & \textbf{28.86\%} & \textbf{58.64\%} & \textbf{63.18\%} & \textbf{88.00\%} & \underline{54.37\%}\\
    Baseline & 76.41\% & \textbf{57.87\%} & 23.54\% & 54.10\% & 60.19\% & 71.00\% & 24.99\%\\ 
    OpenSora-v1-2 & \underline{79.23\%} & 47.22\% & \underline{27.07\%} & 56.18\% & 60.94\% & 85.80\% & \textbf{67.51\%}\\
    OpenSoraPlan-v1-1 & 78.00\% & 47.72\% & 26.52\% & \underline{56.85\%} & \underline{62.28\%} & \underline{86.80\%} & 53.11\%\\
    \bottomrule
  \end{tabular}}
\end{table*}

\noindent \textbf{SP workflow.} We evaluated our SP workflow on a training node with 8 devices, with input size of 32×1080×768 and 128×1080×768, respectively. We take a plain SP implementation as the baseline with replicated VAE and patchifying and equal division of the multi-model sequence (concatenated video and text sequence). Results in Fig.\ref{fig:sp-evaluate} show that separate sharding of the multimodal sequence brings up to 20\% performance gain and frame-wise DP-VAE and patchifying improves training throughput by another 30\%. \\[1ex]
\noindent \textbf{Model Architecture.} Starting from the vanilla baseline of our model, which includes only temporal and spatial attention, we trained both the baseline and the full Vchitect-2.0 model on the same dataset and evaluated their performance using VBench~\cite{vbench}. The results indicate that our model, with the addition of the full-sequence cross-attention module, achieves higher scores in \textit{Total Score}, \textit{Overall Consistency}, \textit{Aesthetic Quality}, and \textit{Imaging Quality}, demonstrating the effectiveness of this design in enhancing semantic motion understanding. This improvement highlights the importance of incorporating full-sequence cross-attention for better video generation performance.

%% file: sec/5_conclusion.tex
\section{Discussion }

While Vchitect-2.0 represents a significant advancement in text-to-video (T2V) generation, there are areas where our models could be improved. The model's performance may decline with longer video sequences due to the accumulation of errors over time, leading to potential temporal inconsistencies. This challenge is common in video diffusion models, as maintaining coherence across extended sequences remains complex.

Besides, the reliance on the Vchitect T2V DataVerse introduces constraints related to data diversity and quality. The dataset may not encompass the full spectrum of real-world scenarios, potentially limiting the model's generalization capabilities. This issue aligns with observations in the field, where models trained on limited datasets struggle to generate high-quality videos across diverse contexts.

Additionally, despite the implementation of memory optimization techniques, Vchitect-2.0's computational demands remain substantial. Training and deploying the model require significant hardware resources, which may not be accessible to all researchers or practitioners. This limitation highlights the need for further research into more efficient architectures or training methods to democratize access to high-quality T2V generation.

In summary, while Vchitect-2.0 marks a substantial step forward in T2V generation, addressing its limitations regarding token length, data diversity, and computational efficiency will be crucial for its broader applicability and effectiveness in diverse real-world scenarios. 

\section{Conclusion}
In this paper, we introduce Vchitect-2.0, a novel parallel transformer architecture designed to address the scaling challenges of video diffusion models. By incorporating a multimodal diffusion block and a sequence-parallel training framework, our approach effectively maintains both spatial and temporal coherence, enabling the generation of high-fidelity videos aligned with text descriptions. The integration of memory-efficient strategies, including activation offloading and optimized inter-node communication, ensures scalability across large-scale distributed systems, even for long video sequences. Our enhanced data processing pipeline further contributes to the performance of Vchitect-2.0 by curating high-quality datasets through rigorous annotation and aesthetic evaluation, leading to better alignment between textual inputs and generated video content. Extensive experiments and evaluations demonstrate that Vchitect-2.0 outperforms existing methods in terms of video quality, training efficiency, and scalability, setting a new benchmark in text-to-video generation. The proposed framework not only advances the state-of-the-art but also lays a strong foundation for future research in scalable video generation. We anticipate that the methods and insights presented here will inspire further exploration and applications in multimodal generation tasks.